\documentclass[a4paper, 12pt]{report}
\usepackage[utf8]{inputenc}
\usepackage[english,italian]{babel}
\usepackage{textcomp}
\usepackage{steinmetz}
\usepackage{url}
\usepackage{bbm}
\usepackage{booktabs}     
\usepackage{color}
\usepackage{indentfirst}  
\usepackage{graphicx}
\usepackage{subfigure}
\usepackage{pdfpages}
\usepackage{subfigure}
\usepackage{wrapfig}
\usepackage{siunitx}      
\usepackage{eurosym}      
\usepackage{enumitem}     
\usepackage{algorithmic}  
\usepackage{algorithm}    
\usepackage{listings}     
\usepackage{pifont}       
\usepackage{geometry}
\usepackage{setspace}     
\usepackage{shapepar}     
\usepackage{amsmath}      
\usepackage{verbatim}
\usepackage{float}
\usepackage{siunitx}
\usepackage{titling}

\usepackage[printonlyused]{acronym}
\usepackage[autostyle,italian=guillemets]{csquotes}
\usepackage[backend=biber, style=numeric, sorting=none, maxbibnames=99]{biblatex}
\usepackage{graphicx}
\usepackage{amssymb}
\usepackage{bm}
\usepackage[normalem]{ulem}
\usepackage[bottom]{footmisc}
\usepackage{dblfloatfix}
\usepackage{tablefootnote}
\makeatletter
 \newcommand{\itemlabelfootnotetext}{\tfn@tablefootnoteprintout \gdef\tfn@fnt{0}}
\makeatother
\definecolor{mygreen}{rgb}{0,0.6,0}

\onehalfspace 

\pretitle{\begin{center}
		\begin{figure}[t]
			\centering
			\includegraphics[width=0.9\linewidth]{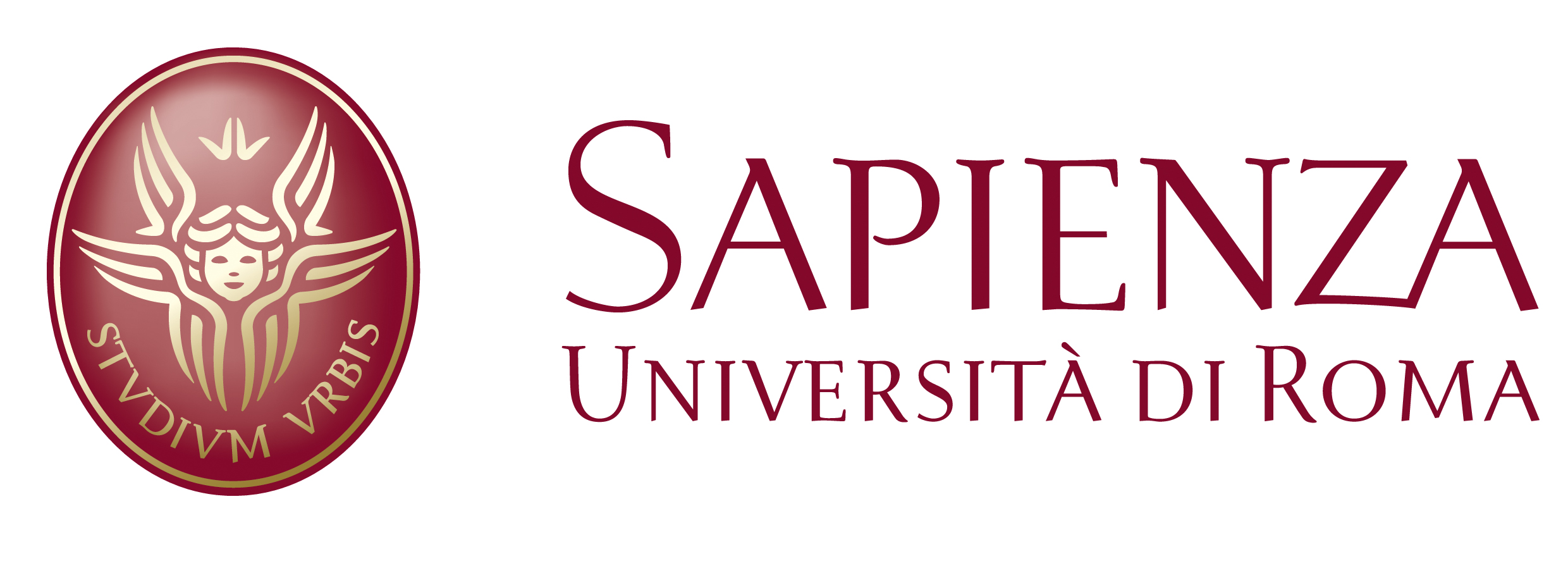}
	\end{figure}}
	
	\title{{\huge Multimodal Deep Domain Adaptation} \\
		\bigskip }
	
	\posttitle{Candidate: Silvia Bucci \hspace{1.5cm} \\Thesis Advisor: Barbara Caputo\\ \hspace{6.1cm} \\Thesis External Advisor: \\Mohammad Reza Loghmani\\
		\bigskip 
		Facoltà di Ingegneria dell'Informazione, Informatica e Statistica\\
		Department of Computer, Control, and Management Engineering\\
		Master of Science in Artificial Intelligence and Robotics\\
		Sapienza University of Rome\end{center}}

\date{A.A. 2017/2018}

\begin{document}
\selectlanguage{english}
\maketitle
\includepdf[pagecommand={\thispagestyle{empty}}]{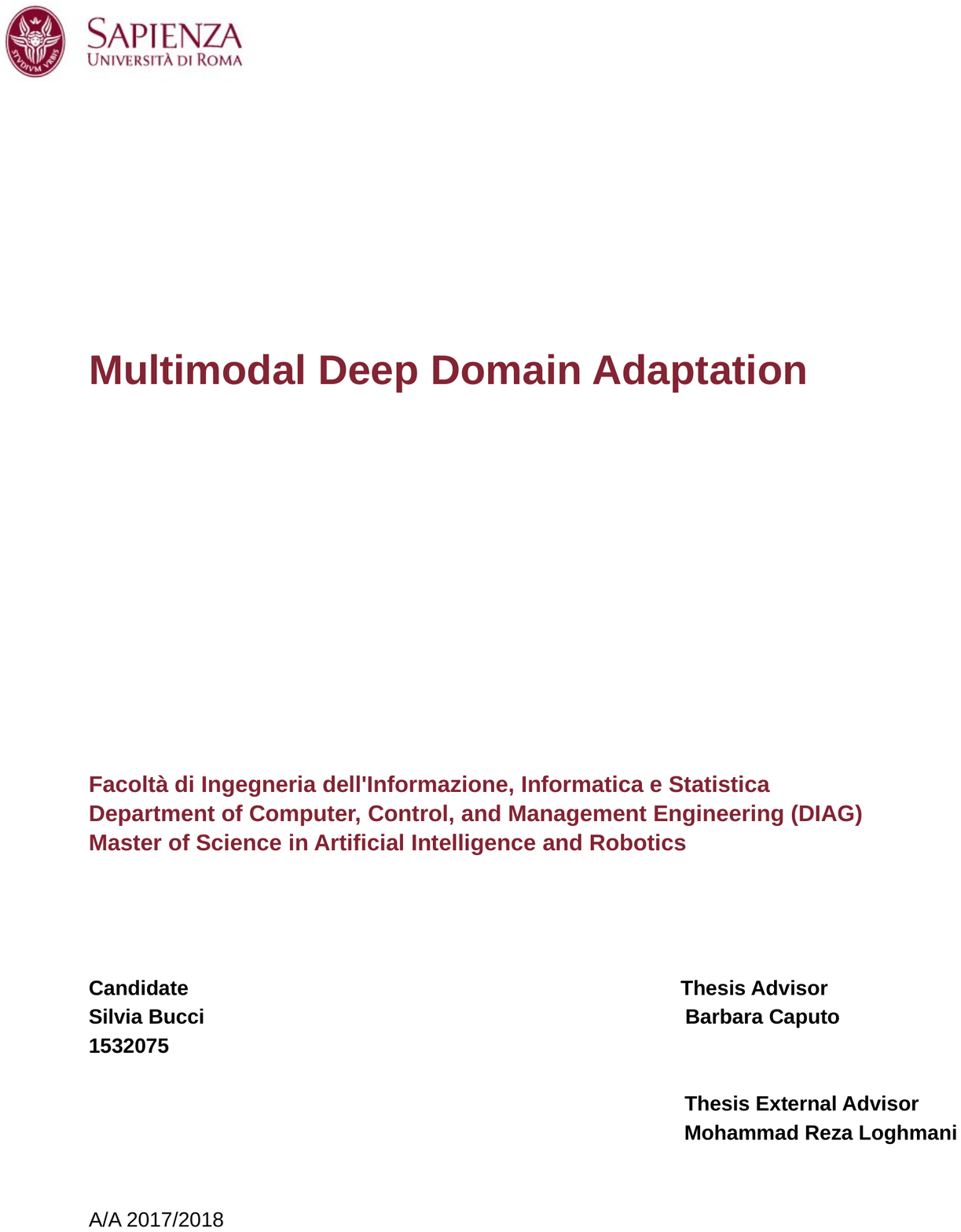}

\cleardoublepage
\thispagestyle{empty}
\centerline{\large\bfseries Ringraziamenti }
\nobreak
\begingroup\small
\noindent
\newline
Ringrazio la mia relatrice, la professoressa Barbara Caputo, per la bellissima opportunità di tesi che mi ha offerto; ringrazio Mohammad per i suoi insegnamenti e per la dedizione con cui mi ha seguito.\newline
Un grande, enorme, infinito GRAZIE va ai miei genitori. Papà, nei tuoi occhi vedo l'amore, nei tuoi occhi mi vedo perfetta, grazie per avermi sempre fatta sentire così, grazie perché quando mi sento persa il tuo sguardo mi fa ritrovare me stessa. Mamma, nei tuoi abbracci trovo tutto quello di cui ho bisogno, il "profumo di mamma" come dicevo da piccola, non hai idea della forza che mi trasmetti con il tuo coraggio, è grazie a te se adesso sono pronta per "correre verso il mondo". Non smetterò mai, mai di ringraziarvi per il vostro amore incondizionato, siete la mia roccia, siete tutto.\newline
Altri due grazie vanno alle mie sorelle, alle mie compagne di gioco preferite da piccole e alle mie compagne di litigate preferite da grandi. La mia dolce Francesca, fragile e sensibile. Grazie, perchè spesso mi sai capire come nessun'altro. Ovunque andrò porterò sempre con me il tuo sguardo in cui vedo un abisso, non smettere MAI di lottare. 
Nicoletta, grazie per tutte le volte in cui ci sei stata, ne abbiamo passate tante, troppe, in questi cinque anni. Ma al termine di questo viaggio posso dire che ne è valsa la pena, sono felice di aver condiviso questa avventura con te.
GRAZIE ai miei quattro pilastri indiscussi, siete il mio tesoro più grande.\newline
Un grazie va a mio nonno che "mi protegge sulla stella più bella che c'è", grazie nonno perchè hai fatto tanto in vita e continui a fare tanto nei miei ricordi. Ringrazio mia nonna che con la sua dolce ingenuitá mi fa ricordare il valore delle cose semplici, sei il caffè e latte che mi preparavi nelle mattine d'estate, sei la focaccia che mi aspettava sempre quando venivo a casa tua, sei una boccata d'aria in questo mondo che non aspetta. Ringrazio mia zia che piano piano sto ritrovando, grazie a te e a Gianni per sostenermi, incoraggiarmi e condividere con me momenti importanti, con la speranza che il nostro rapporto continui a crescere sempre di più. \newline
E poi ci sono i cuginetti che, in realtà, sono dei fratelli. Gae, animo gentile, quando siamo insieme sento che mi sai leggere esattamente per quella che sono, ti ringrazio perchè i tuoi occhi premurosi mi convincono che, nonostante tutto, vado bene così. Chiaretta, ti ringrazio per le nostre intense maratone di studio, ti ringrazio per le avventure in giro per il mondo, ti ringrazio per le chiacchierate infinite dalle quali esco sempre arricchita. Grazie ad entrambi perchè fate parte delle mie certezze. Così come la mia bella zietta lottatrice. Nonostante tutte le sfide che hai dovuto affrontare nella tua vita, mi hai sempre trasmesso serenità. Pensando a te mi ricordo che c'è sempre un motivo per cui vale la pena essere felici. Un grazie va anche a mio zio, che, a suo modo, mi è stato vicino.\newline
Mirco, sei il mio migliore amico, il mio complice perfetto, sei la mia metà. Da quando ho te nella mia vita tutto è più bello. Grazie per la dolcezza con cui, anche senza che io me ne accorga e forse, a volte, in modi che non comprendo, ti prendi cura di me. Ti ringrazio perchè vuoi la mia felicità, perchè non passa un giorno in cui non mi fai sentire amata, non passa un giorno in cui non fai di tutto per farmi sorridere. Le nostre anime sono talmente vicine che a volte sono spaventata. Perchè se perdo te perdo me. Ti ringrazio perchè ogni giorno, sempre di più, sento che con te ho vinto.\newline
Ringrazio la mia persona, Giulia. Spero di darti indietro almeno la metà di quello che dai tu a me. Grazie per farmi capire le cose con umiltà e sincera bontà, sei il mio porto sicuro. Ringrazio Valeriuccia, perchè, seppure così diversa da me, mi ha conquistata. Grazie per la tua spontaneità, per l'affetto che mi dimostri, per i momenti difficili in cui mi sei stata vicina, non li dimenticherò. \newline   
E poi c'è la mia Ariù. Grazie, perchè con la tua profondità e la tua intelligenza mi fai sempre vedere le cose da una prospettiva diversa. Sei una persona speciale, che mi fa sentire speciale. Ti ringrazio per la fiducia che hai sempre avuto in me, ti ringrazio per farmi vedere la vita con la tua ironia, sei uno spasso.\newline
Ringrazio le mie amiche di sempre, Grazia Ludovica Giulia e Francesca. Grazie per avermi fatto conoscere il valore dell'amicizia, siamo cresciute insieme, il nostro legame è cresciuto con noi e per me continuate ad essere un punto di riferimento essenziale.\newline 
E infine un enorme grazie va alla vera ricchezza che mi ha lasciato questa laurea. Grazie al gruppo del "c'è", in particolare grazie a Fra che ormai è una star, grazie a Simone per tutti i gelati che abbiamo divorato dopo le lezioni di probabilità, grazie a Roberto che vedrò sempre come il coniglietto di Galbusera, grazie a Moreno il mio cognatino che solo Dio sa quante ne ha dovute sentire di urla in casa e grazie a Manuel perchè i nostri dialoghi sembrano la trama di un fumetto. Grazie a voi per aver condiviso con me le prime ansie, i primi esami, le prime risate.\newline 
Ringrazio Robertino, perchè grazie a lui mi sono rimasti grossomodo sei anni di vita, grazie Cip per aver seguito il mio consiglio sulla pista di pattinaggio, ringrazio Franci (C.) per le costruttive chiacchierate durante le lezioni di Fisica, ringrazio Scarselli per avermi fatto comprare l'utilissimo manuale di Java e infine un grazie va a Bissolino per avermi mostrato come un essere umano possa vivere dentro una BOLLA (cit.).\newline
Poi c'è il gruppo di matti che si sono aggiunti nella seconda parte di questa avventura. Grazie Bea per essere stata sempre al mio fianco nell'ansia, grazie Darietto per essere riuscito a farmi ridere anche nella più buia disperazione (causata da un certo gabbiano), grazie Atif perchè è solo grazie a te se ho tanti bei piccioni in galleria, ringrazio Stefano per avermi voluto conoscere nonostante le coccinelle, grazie Cate per le giornate indimenticabili al mare, un grazie infinito va a Andrea per avermi fatto conoscere Franco, grazie Niki per le torte che ci hai preparato con amore ad ogni compleanno, ringrazio Dani per il simpaticissimo scherzo al ginocchio e un grazie va a Lorenzo per farmi sempre sorridere con la sua acuta ironia.\newline
Ringrazio tutti per le serate a san Lorenzo, per avermi fatto vivere la magia di questi anni. Porterò con me il ricordo di ognuno di voi.

\textit{La felicità è reale solo se condivisa.}

..perchè alla fine i ringraziamenti sono la parte migliore.

\par\endgroup
\vspace{\fill}
\clearpage
\begin{abstract}
Typically a classifier trained on a given dataset (\textit{source domain}) does not performs well if it is tested on data acquired in a different setting (\textit{target domain}). This is the problem that \textit{domain adaptation} (DA) tries to overcome and, while it is a well explored topic in computer vision, it is largely ignored in robotic vision where usually visual classification methods are trained and tested in the same domain. Robots should be able to deal with unknown environments, recognize objects and use them in the correct way, so it is important to explore the domain adaptation scenario also in this context.

The goal of the project is to define a benchmark and a protocol for multimodal domain adaptation that is valuable for the robot vision community. With this purpose some of the \hbox{state-of-the-art} DA methods are selected: \textit{Deep Adaptation Network} (DAN), \textit{Domain Adversarial Training of Neural Network} (DANN), \textit{Automatic Domain Alignment Layers} (AutoDIAL) and \textit{Adversarial Discriminative Domain Adaptation} (ADDA). Evaluations have been done using different data types: RGB only, depth only and RGB-D over the following datasets, designed for the robotic community: \textit{RGB-D Object Dataset} (ROD), \textit{Web Object Dataset} (WOD), \textit{Autonomous Robot Indoor Dataset} (ARID), \textit{Big Berkeley Instance Recognition Dataset} (BigBIRD) and \textit{Active Vision Dataset}.

Although progresses have been made on the formulation of effective adaptation algorithms and more realistic object datasets are available, the results obtained show that, training a sufficiently good object classifier, especially in the domain adaptation scenario, is still an unsolved problem. Also the best way to combine depth with RGB informations to improve the performance is a point that needs to be investigated more.  
	
\end{abstract}
\tableofcontents
\listoffigures
\listoftables
\chapter*{List of acronyms}
\begin{acronym}
\acro{AI}{Artificial Intelligence}
\acro{CV}{Computer Vision}
\acro{SR}{Service Robots}
\acro{DA}{Domain Adaptation}
\acro{DAN}{Deep Adaptation Network}
\acro{DANN}{Domain Adversarial Training of Neural Network}
\acro{AutoDIAL}{Automatic Domain Alignment Layers}
\acro{ADDA}{Adversarial Discriminative Domain Adaptation}
\acro{ROD}{RGB-D Object Dataset}
\acro{WOD}{Web Object Dataset}
\acro{ARID}{Autonomous Robot Indoor Dataset}
\acro{BigBIRD}{Big Berkeley Instance Recognition Dataset}
\acro{H}{Hilbert Space}
\acro{RKHS}{Reproducing Kernel Hilbert Space}
\acro{MMD}{Maximum Mean Discrepancy}
\acro{MK-MMD}{Multiple Kernel Maximum Mean Discrepancy}
\acro{GRL}{Gradient Reversal Layer}
\acro{CNN}{Convolutional Neural Network}
\acro{QP}{Quadratic Program}
\acro{DA-layers}{Domain Alignment Layers}
\acro{GAN}{Generative Adversarial Network}
\acro{SVM}{Support Vector Machine}
\acro{NLP}{Natural Language Processing}
\acro{SGD}{Stochastic Gradient Descent}
\acro{DDC}{Deep Domain Confusion}

\end{acronym}

\chapter{Introduction}

\begin{figure}[h]
\centering
\includegraphics[width=0.25\textwidth]{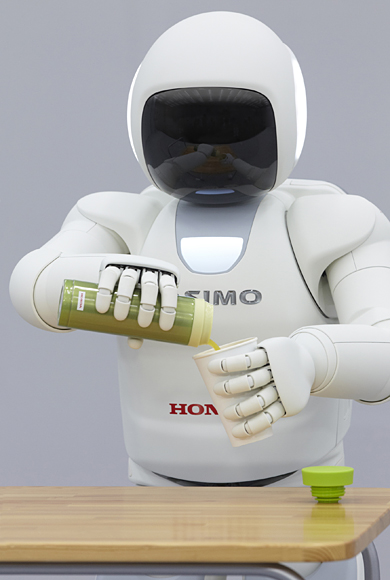}\quad\includegraphics[width=0.63\textwidth]{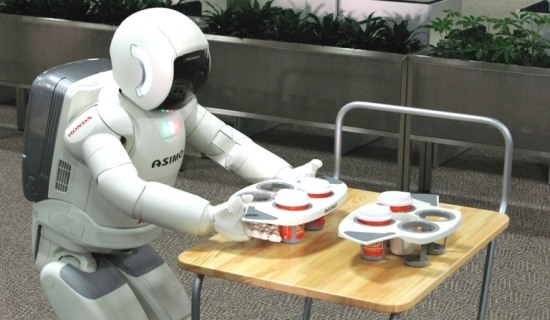}
\caption{\label{fig:asimo} Example of a service robot \cite{asimo1,asimo2}.}
\end{figure}

Object recognition is the ability to re-identify a previously seen object or acknowledge it as belonging to a specific class of objects. In humans, recognition is performed with little effort even when the object appears in different shapes, colors, texture, or if it is partially occluded or seen from a different perspective. One of the open problems in \ac{AI}, and \ac{CV} in particular, is to emulate this skill in artificial systems. Object recognition is particularly important for \ac{SR} that aim at assisting humans in their own environments (e.g. houses and offices). These actions are typically performed in indoor environments as houses or offices. A robot should be able to move in these places full of objects and use them in the proper way to perform its task (see Figure \ref{fig:asimo}).

\begin{figure} [h]
\centering
\includegraphics[width=0.8\textwidth]{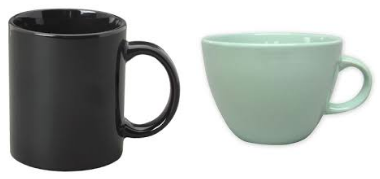}
\caption{\label{fig:coffeemug} A coffee mug can have different characteristics \cite{coffee1,coffee2}.}
\end{figure}

\begin{figure}[h]
\centering
\includegraphics[width=1\textwidth]{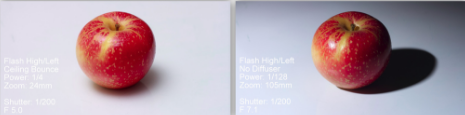}
\caption{\label{fig:apple1} The same object with different illuminations \cite{apple}.}
\end{figure}

Object recognition presents several challenges. Two object instances of the same category can appear very different from each other (see Figure \ref{fig:coffeemug}). Formally, this concept is referred to as intra-class variability, i.e. diversity between samples of the same class. In addition, pictures of the same object can have a different appearance due to environmental conditions such as changes in lighting and viewpoint, variations in scale and background, clutter, and occlusion (see Figure \ref{fig:apple1}).
Usually, in machine learning, the training set is acquired in laboratory under ideal conditions (static lighting, fixed background, same viewpoint, etc.). This gives rise to a set of too "clean" images. As a result, when the classifier is applied to a realistic context, all the problems listed above arise. We can therefore consider training and test sets as if they were two different domains. The discipline that attempts to reduce the discrepancy between images belonging to different domains is \ac{DA}. Typically, the adaptation takes place between a pair of different domains: a \textit{source domain} and a \textit{target domain}.
This project focuses on \textit{unsupervised domain adaptation} so the source domain, with which the classifier is trained, is composed by \hbox{image-labeled} data, while, the target domain, on which the classifier is tested, is made by unlabeled images. The unlabeled condition imposed on the target domain is fundamental because the need to apply domain adaptation algorithms rises mostly from the necessity to deal with unknown environments.
This project regards the evaluation of standard \hbox{state-of-the-art} \ac{DA} algorithms in an unsupervised way on datasets created \hbox{ad-hoc} for service robot functionalities. 
DA algorithms used can be divided in two groups:
\begin{itemize}
\item \textit{Probability distributions alignment} performed minimizing the distance between source and target distributions. This optimization problem can be solved by adding a proper term to the loss function as in \ac{DAN} \cite{DAN}, or using some additional layers that perform a cross-domain adaptation as in \ac{AutoDIAL} \cite{autoDIAL}.
\item \textit{Adversarial alignment} accomplished training adversarial networks. The two networks, one trained for standard object classification and the other trained for domain recognition, can be optimized in the same process as in \ac{DANN} \cite{DANN} or in more phases as in \ac{ADDA} \cite{ADDA}.
\end{itemize}

\begin{figure}[h]
\centering
\includegraphics[width=0.7\textwidth]{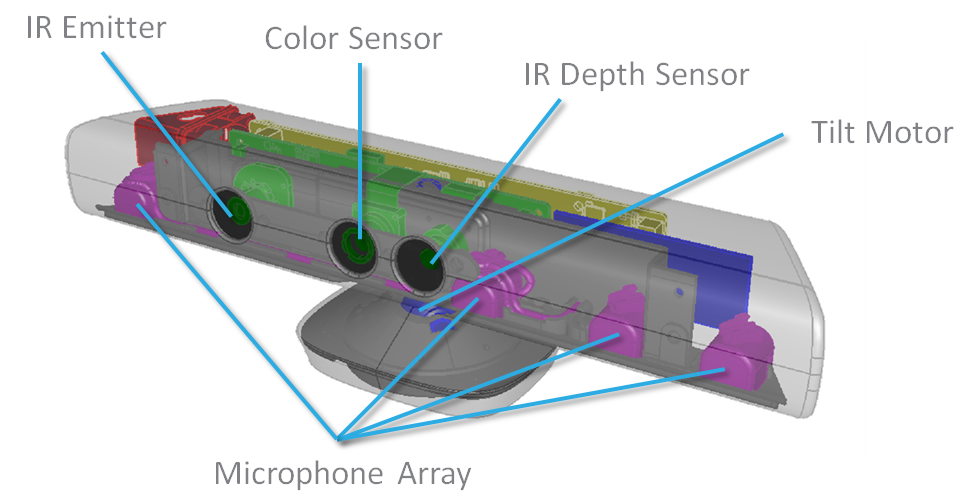}
\caption{\label{fig:kinect} An example of RGB-D camera with sensors description \cite{kinect}.}
\end{figure}

With the advent of Microsoft Kinect \cite{zhang2012microsoft} (see Figure \ref{fig:kinect}), RGB-D camera have become increasingly popular.
These sensors use range imaging technology to provide, in addition to the RGB image, geometric information in form of depth images. Depth images can be very valuable in object recognition since depth information is invariant with respect to some environmental conditions, such as changes in lighting and color. Particularly in robotics, where agents move and operate in the environment, depth is a visual information as important as RGB. 

The goal of this project is to define a benchmark in the DA scenario that is relevant especially for the robot community. The purpose of this research is to investigate the effectiveness of \ac{DA} algorithms in function of the particular input modality chosen. The adaptation is performed with respect to RGB, depth and RGB-D input data. It is important to point out that, to exploit the same networks designed for RGB also with depth informations, it
is necessary to simulate a sort of colorization for them. In the case of RGB-D input modality, instead, a completely different approach, described in the next chapters, is applied.
The datasets chosen for the experiments are compatible with several robotic environments being composed by objects of every day life with which a robot could be faced with. They are: \ac{ROD} \cite{conf/icra/LaiBRF11} \ac{WOD}  \cite{Arid}, \ac{ARID} \cite{Arid}, \ac{BigBIRD} \cite{BigBird} and Active Vision Dataset \cite{ActiveVision}. Analyzing the results, the intention is to identify open issues, specific of robot vision, that current DA approaches fail to solve.

The lower improvements obtained applying DA algorithms on depth images with respect to those obtained on RGB ones leads to the natural conclusion that the algorithms used for this benchmark are more effective on RGB data. Moreover, the use of RGB and depth informations together does not produce the expected results; actually, in some cases, this combination leads to an even worse accuracy with respect to RGB only experiments. So, the optimal method to use depth informations alone, but also the most effective way to exploit depth together with RGB data are still challenges in DA.

The document is composed by a theoretical part followed by an analytical part. In the first one all the algorithms and methods employed for the experiments are described, in the second one the results, with related observations, are shown. In detail, the organization is as follow: in the next chapter some related works are discussed; chapters 3, 4 and 5 show respectively how each \ac{DA} algorithm works, the colorization methods used and the approach adopted for RGB-D experiments; in chapter 6 there is a presentation of the datasets used together with the commented results obtained from the experiments and, finally, in chapter 7 there is a discussion about the findings of the experiments and the conclusions we have drawn.

\chapter{Related works}

During the years, a considerable part of the literature, in the context of \ac{DA}, has focused on the use of linear classifiers \cite{blitzer2006domain,bruzzone2010domain,germain2013pac}. In the last few years, instead, research started to moving towards neural network non-linear representations. For the unsupervised \ac{DA} (the focus of this work) all the approaches can be divided into two groups. The first one refers to the \textit{instance re-weighting}. 
It acts by assigning to each source sample a different weight in function of its similarity with the target domain. Once that the re-weighting is performed, the algorithm trains a classifier, using the re-weighted source domain, that should work well for target data. An example is \cite{zeng2014deep} that introduces an unsupervised domain adaptation algorithm for pedestrian detection that exploits deep autoencoders to weight source data. 
The second group of unsupervised \ac{DA} is composed by all the algorithms that accomplish \textit{features alignment}. This method acts reducing the distance between the source and target domain feature distributions. It can be performed learning \textit{shallow representation models} \cite{gong2012geodesic, long2013transfer} or \textit{deep features} \cite{tzeng2015simultaneous, ganin2014unsupervised}. Due to the task-specific variability of the shallow representations, the deep approach is preferred. This last, in turn, can be accomplished in two ways. 
The first consists in the alignment of source and target distributions either with the insertion of \ac{DA-layers} in the deep network \cite{autoDIAL} or adding a proper term to the \textit{loss function} (a common choice is to use the \ac{MMD} as measure of the dissimilarity between source and target distributions as in \cite{quinonero2008covariate,long2016unsupervised, sun2016return}). The second way to perform features alignment is maximizing a \textit{domain confusion loss} in such a way that the classifier is not able to distinguish between source and target domains
\cite{tzeng2015simultaneous,goodfellow2014generative}. 

Object recognition, together with \ac{DA}, is particularly relevant in robotic applications. One of the main problem in this field is the lack of proper datasets to train robust classifiers that work well in realistic contexts (such as robotic environments). The idea behind some previous works is the use of data from the World Wide Web as source domain properly adapted to robotic purposes. In particular, \cite{lai20093d} and \cite{lai2010object} use objects from Google’s 3D Warehouse \cite{google} as source domain. To overcome the differences between web data and real data, a small set of labeled point clouds recorded by mobile robots in realistic environments is added to source domain and domain adaptation algorithms are applied.

Most of the literature both in computer and robotic vision field is focused on RGB images, several domain adaptation benchmarks have been defined using this type of information. Some examples can be found in works in which new algorithms are proposed: in these types of publications a benchmark is necessary to verify the performance of the new domain adaptation algorithm designed (some examples are \cite{DAN,DANN,autoDIAL,ADDA,tzeng2014deep,long2016unsupervised}). Depth, on the other hand, is a kind of information rarely considered in the creation of domain adaptation benchmarks despite this data is particularly relevant in robotics where the aim is to model agents that should be able to move and act in the environment.
Even more rare is the use of the RGB-D informations in these scenarios. Both these modalities have been investigated by really few works, as in \cite{novi}. For sure, one of the reasons of this gap is in the lack of proper datasets in which RGB and depth informations can be used together. 

The contribution of this work is to define a benchmark for the robot vision community using RGB, depth and RGB-D modalities as input data using some state-of-the-art domain adaptation algorithms.

\chapter{Domain Adaptation Algorithms}

In this chapter we revise into details the domain adaptation algorithms used in this thesis. We explicitly focused on the unsupervised domain adaptation scenario, characterized by a \textit{source}  domain  $D_s= \{(x^s_{i},y^s_{i})\}^{n_s}_{i=1}$ with $n_{s}$ labeled examples, and a \textit{target} domain $D_t= \{(x^t_{j})\}^{n_t}_{j=1}$ with $n_t$ unlabeled examples.

\section{Deep Adaptation Network}

Recent studies demonstrate that the features transferability decreases significantly in the higher layers of a deep network with an increment of domain discrepancy. Practically, the features computed in the higher layers depend a lot on the specific dataset and task. The goal of \ac{DAN} is to increase the transferability in these task-specific layers by generalizing deep convolutional neural networks to the domain adaptation scenario. To reach this goal, the hidden representations of all these task-specific layers are embedded to a \ac{RKHS} where the mean embeddings of various domain distributions can be matched. Because mean embedding matching is affected by the kernel choices, an optimal multi-kernel selection procedure is designed to reduce further the domain discrepancy.

\subsection{Multiple Kernel Maximum Mean Discrepancy} 

The \ac{H} is the space of the features in which the data are mapped to make them linearly separable in the classification. The \ac{RKHS} is composed by the inner products of the elements of \ac{H} and it is defined by a type of kernel.
The \ac{MMD} expresses the "distance" between the images of the source domain and the images of the target domain. This quantity is computed in a \ac{RKHS} between the mean embedding ($\mu_p$) of p (probability distribution of the source domain) and the mean embedding ($\mu_q$) of q (probability distribution of the target domain). 
So the \ac{RKHS} distance between two probability distributions will be the difference of the square root between two elements of \ac{RKHS} (two inner products):
\begin{equation}
d^2(p,q)  \triangleq \|  \textbf{E}_p[\phi(\textbf{x}^s)] - \textbf{E}_q[\phi(\textbf{x}^t)] \|^2_{H_k}
 \label{MKMMD}
\end{equation}

\[where \]

\[ \textbf{E}_p[f(\textbf{x})] = < f(\textbf{x}), \mu_k(p)>  \forall f \in H_k \]
\[ \textbf{E}_q[f(\textbf{x})] = < f(\textbf{x}), \mu_k(q)>  \forall f \in H_k \]
\\
Notice that $p=q$ iff $d^2_k(p,q)=0$.
The method is based on the multiple kernel variant of \ac{MK-MMD}: the characteristic kernel used to compute the inner product associated to the feature map $\phi$ ($k(\textbf{x}^s,\textbf{x}^t) = <\phi(\textbf{x}^s),\phi(\textbf{x}^t)>$ is not fixed a priori but it is given as the convex combination of $m$ positive semi-definite kernels $\{k_u\}$:

\[  K \triangleq \left\{ k = \sum_{u=1}^{m}{ \beta_u k_u } : \sum_{u=1}^{m}{ \beta_u = 1}, \beta_u \geqslant 0, \forall u   \right\}
\]
where the positivity constraints on coefficients $\{\beta\}$ are used to ensure that the derived multi-kernel $k$ is characteristic. The kernel selected to compute the mean embeddings of \textit{p} and \textit{q} is important to guarantee the test power and low test error.

\subsection{Model}

\begin{figure}
\centering
\includegraphics[width=0.8\textwidth]{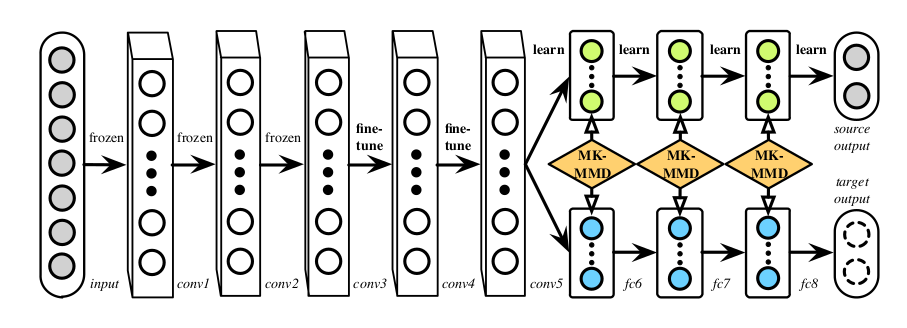}
\caption{\label{fig:DAN}The Deep Adaptation Network architecture for learning transferable features \cite{DAN}.}
\end{figure}

The starting point of this method is a deep \ac{CNN} but, just adapting \ac{CNN} via fine-tuning to the target domain (that has no labels) is difficult and could lead to over-fitting. Therefore the idea is to model a deep adaptation network (\ac{DAN}) that can take advantage from both source domain (with labeled data) and target domain (with unlabeled data) (see Figure \ref{fig:DAN}). 
In a classical \ac{CNN} each $f_c$ layer $\ell$ acquires a nonlinear mapping $\textbf{h}^{\ell}_i = f^{\ell} (\textbf{W}^{\ell}\textbf{h}^{\ell-1}_i + \textbf{b}^{\ell}) $ where $\textbf{h}^{\ell}_i$ is the $\ell$th layer hidden representation of point $\textbf{x}_i$, $\textbf{W}^{\ell}$ and $\textbf{b}^{\ell}$ are the weights and bias of the $\ell$th layer, and $f^{\ell}$ is the activation function, taking $f^{\ell}(\textbf{x}) = max(\textbf{0},\textbf{x})$ (rectifiers units) for hidden layers or $f^{\ell}(\textbf{x}) = e^\textbf{x} / \sum_{j=1}^{|\textbf{x}|}e^{x_j}$ (softmax units) for the output layer. Since $\Theta = \{\textbf{W}^{\ell},\textbf{b}^{\ell}\}^{l}_{\ell = 1}$ are the \ac{CNN} parameters, the empirical risk of \ac{CNN} will be:

\[    \min_{\Theta}\frac{1}{n_a} \sum_{i=1}^{n_a}{ J(\theta(\textbf{x}^a_i),y^a_i)} \]

in which $J$ is the cross-entropy loss function, and $\theta(\textbf{x}^a_i)$ represents the condition probability that the \ac{CNN} will compute a correct prediction $y^a_i$ for the example $\textbf{x}^a_i$.

The novelty of \ac{DAN} is in the contribution of a \textit{MK-MMD-based multi-layer} adaptation regularizer to the \ac{CNN} risk:

\begin{equation}
 \min_{\Theta}\frac{1}{n_a} \sum_{i=1}^{n_a}{ J(\theta(\textbf{x}^a_i),y^a_i)} + \lambda \sum_{\ell=l_1}^{l_2} d^2_k(D^{\ell}_s,D^{\ell}_t)
 \label{DANrisk}
\end{equation}

where $\lambda > 0$ is a parameter of penality, $l_1$ and $l_2$ are the layers between which the adaptation regularizer acts, $D^{\ell}_{*} = \{\textbf{h}^{* \ell}_i \}$ is the hidden representation of the $\ell$th layer for the source and target examples, and $d^2_k(D^{\ell}_s,D^{\ell}_t)$ is the \ac{MK-MMD} among the source and target examples computed in the $\ell$th layer representation. The \ac{MK-MMD} term will lead source and target distributions to become similar under the hidden representations of fully connected layers.

Two significant points that differentiate \ac{DAN} from previous works are:
\begin{itemize}
\item \textit{multi-layer adaptation}. It is not enough to adapt a single layer to reduce sufficiently the dataset bias between source and target domain because there is more than one layer that is not transferable. Furthermore, adapting the representation layers and the classification layer together, we could link the domain discrepancy that characterizes the marginal distribution and the conditional distribution, a critical point for domain adaptation.
\item \textit{multi-kernel adaptation}. The choice of the kernel is important: different kernels embed probability distributions in different \ac{RKHS}s in which different orders of sufficient statistics can be accentuated.
\end{itemize}

\subsection{Algorithm}

\textbf{Learning $\Theta$}. The computation of best parameters $\Theta$ in \eqref{DANrisk} is executed in this way:
\begin{itemize}
\item For \textbf{\ac{MK-MMD}} \eqref{MKMMD} the unbiased estimate is used because it can be calculated with linear complexity. In particular $d^2_k(p,q) = \frac{2}{n_s} \sum_{i=1}^{\frac{n_s}{2}}{g_k(\textbf{z}_i)}$, where $\textbf{z}_i \triangleq (\textbf{x}^s_{2i-1},\textbf{x}^s_{2i},\textbf{x}^t_{2i-1},\textbf{x}^t_{2i})$ and $g_k(\textbf{z}_i) \triangleq k(\textbf{x}^s_{2i-1},\textbf{x}^s_{2i})+k(\textbf{x}^t_{2i-1},\textbf{x}^t_{2i})-k(\textbf{x}^s_{2i-1},\textbf{x}^t_{2i})-k(\textbf{x}_{2i}^s,\textbf{x}_{2i-1}^t)$. This computation is done with cost $O(n)$. To optimize this part of DAN risk we just need to compute the gradients $\frac{\partial g_k(\textbf{z}^{\ell}_i)}{\partial{\Theta^{\ell}}}$ for the quad-tuple $\textbf{z}^{\ell}_i = (\textbf{h}^{s \ell}_{2i-1},\textbf{h}^{s \ell}_{2i},\textbf{h}^{t \ell}_{2i-1},\textbf{h}^{t \ell}_{2i})$ of the $\ell$th layer hidden representation. 

\item Similarly to the gradient of \ac{MK-MMD}, we calculate the corresponding gradient of \textbf{\ac{CNN} risk} $\frac{\partial J(\textbf{z}_i)}{\partial{\Theta^{\ell}}}$, where $J(\textbf{z}_i) = \sum_{i'}{J(\Theta(\textbf{x}^a_{i'}),y^a_{i'})}$ and $\{(\textbf{x}^a_{i'},y^a_{i'})\}$ correspond to the labeled examples in quad-tuple $\textbf{z}_i$. 
\end{itemize}

In summary, to perform mini-batch updates, we calculate the gradient of the objective function \eqref{DANrisk} with respect to the $\ell$th layer parameter $\Theta^{\ell}$ in this way: 

\begin{equation}
\nabla_{\Theta^{\ell}} = \frac{\partial J(\textbf{z}_i)}{\partial{\Theta^{\ell}}} + \lambda \frac{\partial g_k(\textbf{z}^{\ell}_i)}{\partial{\Theta^{\ell}}}
\end{equation}

Since kernel $k$ is the linear combination of $m$ Gaussian kernels $\{ k_u(\textbf{x}_i,\textbf{x}_j) = \exp ^{-\| \textbf{x}_i-\textbf{x}_j\| ^2 / \gamma_u} \}$, the gradient $\frac{\partial g_k(\textbf{z}^{\ell}_i)}{\partial{\Theta^{\ell}}}$ can be quickly solved with the chain rule.\newline

\textbf{Learning $\pmb{\beta}$}. 
For the selection of the optimal kernel parameter $\beta$ in \ac{MK-MMD} it should be optimized this term: 
\begin{equation}
\max_{k \in K}{d^2_k(D^{\ell}_s,D^{\ell}_t)\sigma^{-2}_k}
\label{Opt}
\end{equation}

where $\sigma^2_k =  \textbf{E}_\textbf{z}g^2_k(\textbf{z}) - [\textbf{E}_\textbf{z}g_k(\textbf{z})]^2$ is the estimation variance. \newline 
Being $ \textbf{d} = (d_1,d_2,...,d_m)^T$, each $d_u$ is MMD through kernel $k_u$. Covariance $\textbf{Q} = cov(g_k) \in \Re^{mxm} $ can be calculated in $O(m^2n)$ cost, i.e. $\textbf{Q}_{uu^{'}} = \frac{4}{n_s} 
\sum_{i=1}^{\frac{n_s}{4}}{g_{k_u}^{\nabla}(\overline{\textbf{z}}_i) g_{k_u^{'}}^{\nabla}(\overline{\textbf{z}}_i)} $ where $\overline{\textbf{z}}_i \triangleq (\textbf{z}_{2i-1},\textbf{z}_{2i})$ and $g_{k_u}^{\nabla}(\overline{\textbf{z}}_i) \triangleq  g_{k_u}(\textbf{z}_{2i-1})-g_{k_u}(\textbf{z}_{2i})$. Consequently \eqref{Opt} becomes this simpler \ac{QP},

\begin{equation}
\min_{\textbf{d}^T\beta=1,\beta \ge 0}{\pmb{\beta}^T(\textbf{Q}+\varepsilon \textbf{I})\pmb{\beta}},
\label{Opt2}
\end{equation}

where $\varepsilon = 10^{-3}$ acts as regularizer to transform the problem in a well-defined way. We can observe that the \ac{DAN} risk \eqref{DANrisk} is substantially a minimax problem:  
\begin{equation}
\min_{\Theta} \max_{K} {d^2_k(D^{\ell}_s,D^{\ell}_t)\sigma^{-2}_k}\end{equation} These two operations act to reach an efficient adaptation for the domain discrepancy, with the purpose of consolidating the transferability of \ac{DAN} features.

\newpage
\section{Domain Adversarial Training of Neural Network}

\begin{figure}[h]
\centering
\includegraphics[width=0.8\textwidth]{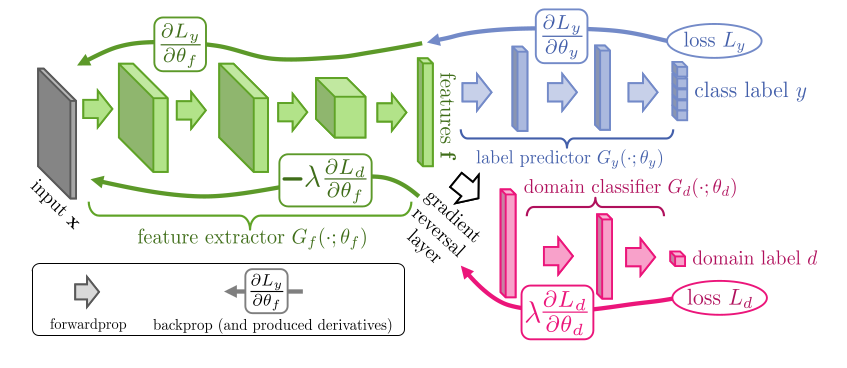}
\caption{\label{fig:DANN}The Domain Adversarial Training of Neural Network architecture with the deep \textit{features extraction} (in green), the deep \textit{label predictor} (in blue) and the \textit{gradient reversal layer} (in pink) \cite{DANN}.}
\end{figure}

The central point of this method is the implementation of a mapping between source and target domain such that the classifier trained on the source works well also if it is tested on the target. The algorithm focuses on learning features that are both discriminative and domain-invariant.
With this in mind two classifiers are optimized:
\begin{itemize}
\item the \textit{label classifier} that is the standard class labels predictor.
\item the \textit{domain classifier} whose intention is to distinguish between the source and target domains.
\end{itemize}
The optimal feature mapping is found by \textit{minimizing} the loss function of the label classifier and by \textit{maximizing} the loss function of the domain classifier. The \textit{maximization} then acts \textit{adversarially} to the domain classifier. The innovation of \ac{DANN} algorithm is in the adding of a \ac{GRL} in the classical \ac{CNN} (Figure \ref{fig:DANN}).

\subsection{H-Divergence}

\textbf{Definition 1 (Ben-David et al., 2006, 2010; Kifer et al., 2004)}
\textit{Given two domain distributions $D^X_S$ and $D^X_T$ over $X$, and a hypothesis class $H$ (that we assume to be a set of binary classifiers $\eta : X \rightarrow \{0,1\}$), the $H$-divergence between $D^X_S$ and $D^X_T$ is}

\begin{equation*}
d_H(D^X_S,D^X_T) = 2 \sup_{\eta \in H} {\left| \Pr_{x \sim D^X_S} [\eta(\textbf{x}) = 1] - \Pr_{x \sim D^X_T} [\eta(\textbf{x}) = 1] \right| }.
\end{equation*}

This quantity is the ability of the hypothesis class $H$ to recognize between examples produced by $D^X_S$ from examples produced by $D^X_T$. An important property is that, if the hypothesis class $H$ is symmetric, the \textit{empirical H-divergence} among two examples $S \sim (D^X_S)^n$ and $T \sim (D^X_T)^{n'}$ can be computed in this way:

\begin{equation}
\hat{d}_H(S,T) = 2 \left( 1- \min_{\eta \in H}{\left[ \frac{1}{n} \sum_{i=1}^{n}{I[\eta(\textbf{x}_i) = 0] + \frac{1}{n'} \sum_{i=n+1}^{N}{I[\eta(\textbf{x}_i) = 1] }}         \right]} \right)
\label{Hdivergence}
\end{equation}

in which $I[a]$ is 1 if \textit{a} is true, and 0 if \textit{a} is false.

\subsection{Model}

For simplicity we analyze the method in a neural network composed by only one hidden layer which learns a function $G_f : X \rightarrow \mathbb{R}^D$ (with $X = \mathbb{R}^m$) that maps each example in a $D$-dimensional representation:

\begin{equation}
G_f(\textbf{x};\textbf{W},\textbf{b}) = \text{sigm}(\textbf{Wx}+\textbf{b})
\label{outputofhiddenlayer}
\end{equation}
\\
where $ \text{sigm}(\textbf{a}) = \left[ \frac{1}{1+exp(-a_i)} \right]^{|\textbf{a}|}_{i=1}$. In the same way, the prediction layer learns a function $G_y : \mathbb{R}^D \rightarrow [0,1]^L $:
\begin{equation*}
G_y(G_f(\textbf{x};\textbf{W},\textbf{b}); \textbf{V}, \textbf{c}) = \text{softmax}(\textbf{V}G_f(\textbf{x} ;\textbf{W},\textbf{b})+\textbf{c})
\end{equation*}

where $\text{softmax}(\textbf{a}) = \left[ \frac{\text{exp}(a_i)}{\sum_{j=1}^{|\textbf{a}|} {\text{exp}(a_j)}} \right]^{|\textbf{a}|}_{i=1} $.

In these formulations $(\textbf{W},\textbf{b}) \in \mathbb{R}^{D \times m} \times \mathbb{R}^D $ are the parameters of the hidden layer $G_f$ and  
$(\textbf{V}, \textbf{c}) \in \mathbb{R}^{L \times D} \times \mathbb{R}^L$ (with $L=|Y|$)  are the parameters of the prediction layer $G_y$. Given a source sample $(\textbf{x}_i,y_i)$ the typical classification loss is:

\begin{equation*}
\mathcal{L}_y(G_y(G_f(\textbf{x};\textbf{W},\textbf{b}); \textbf{V}, \textbf{c}),y_i) = \text{log} \frac{1}{G_y(G_f(\textbf{x};\textbf{W},\textbf{b}); \textbf{V}, \textbf{c})_{y_i}}
\end{equation*}

So during training the optimization problem will be:

\begin{equation}
\min_{\textbf{W},\textbf{b},\textbf{V},\textbf{c}} \left[ \frac{1}{n} \sum_{i=1}^{n} \mathcal{L}_y^i(G_y(G_f(\textbf{x};\textbf{W},\textbf{b}); \textbf{V}, \textbf{c}),y_i) +\lambda \cdot R(\textbf{W},\textbf{b})  \right]
\label{optimization}
\end{equation}
in which $n$ is the number of source examples and $R(\textbf{W},\textbf{b})$ is a regularizer.\newline
The central point of the method is in the addition of a term using the Definition 1 of \textit{H-divergence}, this term is called \textit{domain regularizer}. Consider the output of the hidden layer $G_f(\cdot)$ \eqref{outputofhiddenlayer} for source example:

\begin{equation*}
S(G_f) = \{ G_f(\textbf{x}) | \textbf{x} \in S\}
\end{equation*}

and, for target example:

\begin{equation*}
T(G_f) = \{ G_f(\textbf{x}) | \textbf{x} \in T \}.
\end{equation*}

Using the empirical $H-divergence$ \eqref{Hdivergence} ($H$ is a symmetric hypothesis class) between $S(G_f)$ and $T(G_f)$ we have:

\begin{equation}
\hat{d}_{H}(S(G_f),T(G_f)) = 
\end{equation}
\begin{equation}
2 \left(  1-\min_{\eta \in H} \left[\frac{1}{n} \sum_{i=1}^{n} {I[\eta(G_f(\textbf{x}_i))=0]} + 
\frac{1}{n'} \sum_{i=n+1}^{N} {I[\eta(G_f(\textbf{x}_i))=1] } \right] \right)
\label{nodomainclass}
\end{equation}

We estimate the min part of \eqref{nodomainclass} with a \textit{domain classifier layer} $G_d$ that learns a function (\textit{logic regressor}) $G_d : \mathbb{R}^D \rightarrow [0,1]$ with parameters $(\textbf{u},z) \in \mathbb{R}^D \times \mathbb{R}$. This classifier gives the probability that an example is from source dataset or target dataset.

\begin{equation*}
G_d(G_f(\textbf{x}); \textbf{u},z) = \text{sigm}(\textbf{u}^TG_f(\textbf{x})+z)
\end{equation*}

Its loss is:

\begin{equation*}
\mathcal{L}_d (G_d(G_f(\textbf{x}_i)),d_i) = d_i\text{log}\frac{1}{G_d(G_f(\textbf{x}_i))} + (1-d_i)\text{log}\frac{1}{1-G_d(G_f(\textbf{x}_i))}
\end{equation*}

in which $d_i$ is the binary \textit{domain label} for the $i$-th example ($d_i=0$ if $\textbf{x}_i$ belong to the source domain, $d_i=1$ if $\textbf{x}_i$ belong to the target domain).\newline
We add a domain adaptation term in the equation \eqref{optimization} as regularizer term:

\begin{equation*}
R(\textbf{W},\textbf{b}) = \max_{\textbf{u},z} \left[ -\frac{1}{n} \sum_{i=1}^{n} \mathcal{L}_d^i(\textbf{W},\textbf{b},\textbf{u},z) - \frac{1}{n'} \sum_{i=n+1}^{N} \mathcal{L}_d^i(\textbf{W},\textbf{b},\textbf{u},z)    \right]
\end{equation*}

in which $\mathcal{L}_d^i(\textbf{W},\textbf{b},\textbf{u},z) = \mathcal{L}_d (G_d(G_f(\textbf{x}_i;\textbf{W},\textbf{b}); \textbf{u},z),d_i)$. Since $2(1-R(\textbf{W},\textbf{b}))$ is a surrogate of $\hat{d}_H(S(G_f),T(G_f))$, this term is the approximation of the $H$-divergence equation \eqref{nodomainclass}.

In summary, we can write the final optimization problem of \eqref{optimization} in this way:

\begin{equation*}
E(\textbf{W},\textbf{V},\textbf{b},\textbf{c},\textbf{u},z) = \frac{1}{n} \sum_{i=1}^{n} \mathcal{L}_y^i (\textbf{W},\textbf{b},\textbf{V},\textbf{c}) -
\end{equation*}
\begin{equation*}
\lambda \left( \frac{1}{n} \sum_{i=1}^{n} \mathcal{L}_d^i(\textbf{W},\textbf{b},\textbf{u},z) + \frac{1}{n'} \sum_{i=n+1}^{N} \mathcal{L}_d^i(\textbf{W},\textbf{b},\textbf{u},z) \right)
\end{equation*}

and the optimal parameters are found solving:

\begin{equation*}
(\hat{\textbf{W}},\hat{\textbf{V}},\hat{\textbf{b}},\hat{\textbf{c}}) = \underset{\textbf{W},\textbf{V},\textbf{b},\textbf{c}}{\operatorname{\arg\min}} \  E(\textbf{W},\textbf{V},\textbf{b},\textbf{c},\hat{\textbf{u}},\hat{z})
\end{equation*}

\begin{equation*}
(\hat{\textbf{u}},\hat{z}) = \underset{\textbf{u},z}{\operatorname{\arg\max}} \  E(\hat{\textbf{W}},\hat{\textbf{V}},\hat{\textbf{b}},\hat{\textbf{c}},\textbf{u},z)
\end{equation*}

\newpage
\section{Automatic Domain Alignment Layers}

The technique used in this method aims at aligning the features of the source and target domains to a canonical one with the addition of \textit{\ac{DA-layers}} in the standard networks. The novelty of this approach is in the self-tuning of the degree of feature alignment in function of the different level of the deep network in which the new layers are embedded. This is the real innovation of this method: the dynamism of domain alignment parameters.

\subsection{Source and target predictor}

\begin{figure}[h]
\centering
\includegraphics[width=0.9\textwidth]{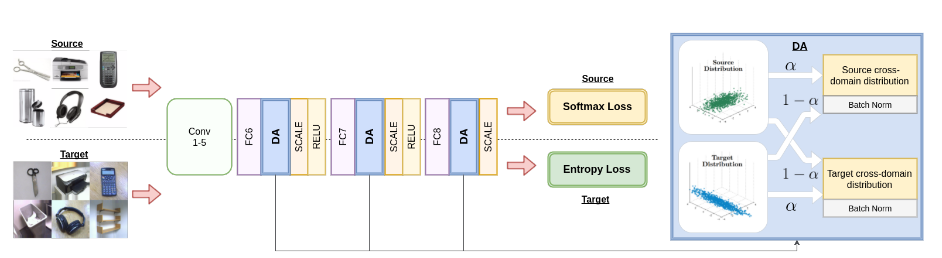}
\caption{\label{fig:AutoDIAL} Automatic Domain Alignment Layers applied on Alexnet. \cite{autoDIAL}}
\end{figure}

The typical assumption that is done by the majority of domain adaptation methods is that the domain alignment between source and target can be done using the same predictor. In this work an impossibility theorem is considered, that states the inability  of a learner to perform the domain adaptation among distributions characterized by a \textit{covariance shift} without additional hypothesis of dependence among them. The \ac{AutoDIAL} method implements two different predictors for the source and the target domain starting from two deep neural networks with the same structure and the same weights. Both networks have also the same number of \ac{DA-layers} in the same positions (Figure \ref{fig:AutoDIAL}). 

Since the source and target distributions are different, the transformations implemented by the adaptation layers to align the two distributions with a reference one are different. Basically the \ac{DA-layers} are Batch Normalization layers but, instead of using the standard implementation with the first and second-order moment computed on the input distribution, a cross-domain bias is used: the second-order moment is contaminated with the other domain.

\subsection{Domain Alignment Layers}

Let $q^s$ be the probability distribution of source input data $x_s$ and $q^t$ the probability distribution of target input data $x_t$, let us assume $q_{\alpha}^{st} = \alpha q^s +(1-\alpha)q^t$ and $q_{\alpha}^{ts} = \alpha q^t +(1-\alpha)q^s$ be cross-domain distributions with the mixed factor $\alpha \in \left[0.5,1\right]$. The outcome of the \ac{DA-layers} will be, for the source network

\begin{equation*}
DA(x_s;\alpha) = \frac{x_s - \mu_{st,\alpha}}{\sqrt{\epsilon + \sigma^2_{st,\alpha}}}
\end{equation*}

and, for the target network

\begin{equation*}
DA(x_t;\alpha) = \frac{x_t - \mu_{ts,\alpha}}{\sqrt{\epsilon + \sigma^2_{ts,\alpha}}}
\end{equation*}

with $\epsilon > 0$ a small number useful in case of variance equal to zero, \hbox{$\mu_{st,\alpha} = \text{E}_{x \sim q_{\alpha}^{st}}[x]$} the mean and  $\sigma^2_{st,\alpha}=\textit{Var}_{x \sim q_{\alpha}^{st}}[x] $ the variance of $x \sim q_{\alpha}^{st}$. Similarly, $\mu_{ts,\alpha} = \text{E}_{x \sim q_{\alpha}^{ts}}[x]$ the mean and  $\sigma^2_{ts,\alpha}=\textit{Var}_{x \sim q_{\alpha}^{ts}}[x] $ the variance of $x \sim q_{\alpha}^{ts}$. It can be noticed that by using $\alpha = 1$ there is an independent alignment of source and target domains: the \ac{DA-layers} calculate two completely different functions for the source and the target predictors. Using instead $\alpha = 0.5$ the new layers give rise to the same function for the predictors of the two domains: they are transformed likewise ($q^{st}_{0.5} = q^{ts}_{0.5}$) and there is no domain alignment. Thus the choice of the mixing factor is crucial, it is not decided a priori but it is learned in the training process.

\subsection{Training}

The common weights of source and target predictors and the mixing factor related to the \ac{DA-layers} are computed during the training phase by employing a source labeled dataset and a target unlabeled dataset. We have the following a posterior distribution of the set of parameters $\theta$ (weights and mixing factor):

\begin{equation}
\pi(\theta|S,T) \propto  \pi(y_s|x_s,T,\theta)\pi(\theta|T,x_s)
\label{postepriorterm}
\end{equation}

with $x_s = \{ x_1^s, \dots , x_n^s \}$ and $y_s = \{ y_1^s, \dots , y_n^s \}$ corresponding to data and labels of the source domain. This quantity is maximized to find the best value for $\theta$:

\begin{equation*}
\hat{\theta} \in  \underset{\theta \in \Theta}{\operatorname{\arg\max}} \  \pi(\theta|S,T)
\end{equation*}

In the expression \eqref{postepriorterm} two terms can be isolated: $\pi(y_s|x_s,T,\theta)$ the likelihood of $\theta$ w.r.t the source domain and $\pi(\theta|T,x_s)$ the prior term determined by the target domain. Truly each of this two terms depends on both domains thanks to the mixing factor $\alpha$. The likelihood term can be rewritten in this way (since the data are $i.i.d.$ for hypothesis):
\begin{equation*}
\pi(y_s|x_s,T,\theta) = \prod^n_{i=1}{f^{\theta}_s(y_i^s;x_i^s)}
\end{equation*}

in which $f^{\theta}_s(y_i^s;x_i^s)$ expresses the probability that the source predictor assigns at the example $x_i^s$ the label $y_i^s$. The prior distribution is computed as a function of the level of label uncertainty (for an hypothesis $\theta$) when a predictor is used on target samples:

\begin{equation*}
h(\theta|T,x_s) = -\frac{1}{m} \sum^m_{i=1} \sum_{y \in Y} f^{\theta}_t(y;x^t_i)\text{log}f^{\theta}_t(y;x^t_i)
\end{equation*}

in which $f^{\theta}_t(y;x^t_i)$ is the probability that the target predictor assigns at the example $x^t_i$ the label $y$. This term is the empirical entropy of $y|\theta$ conditioned on $x$ from which a prior distribution can be derived:

\begin{equation*}
\pi(\theta|T,x_s) \propto \text{exp}(-\lambda h(\theta|T,x_s)) 
\end{equation*}

with the constraint $\int h(\theta|T,x_s)\pi(\theta|T,x_s)s\theta = \varepsilon$ $(\varepsilon > 0)$ expressing how low should be the label uncertainty. The parameter $\lambda$ is the Lagrange multiplier associated with $\varepsilon$. In conclusion the loss function used during the training process is:
\begin{equation}
L(\theta) = L^s(\theta)+\lambda L^t(\theta)
\label{lossautodial}
\end{equation}

where

\begin{equation*}
L^s(\theta) = - \frac{1}{n} \sum_{i=1}^n \text{log} f^{\theta}_s(y^s_i;x^s_i)
\end{equation*}

\begin{equation*}
L^t(\theta) =  -\frac{1}{m} \sum^m_{i=1} \sum_{y \in Y} f^{\theta}_t(y;x^t_i)\text{log}f^{\theta}_t(y;x^t_i)
\end{equation*}

\newpage
\section{Adversarial Discriminative Domain Adaptation}

\begin{figure}[h]
\centering
\includegraphics[width=1\textwidth]{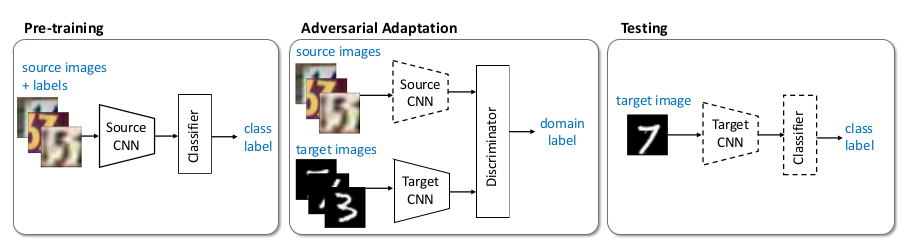}
\caption{\label{fig:ADDA} Phases of the Adversarial Discriminative Domain Adaptation method. Dotted lines denote pinned network parameters \cite{ADDA}.}
\end{figure}

The method is based on a \ac{GAN}s learning where two networks (a generator and a discriminator) are involved: the generator yields images so that confuses the discriminator which attempts to recognize them from real image examples. This mechanism in domain adaptation is used to make sure that the network is not able to distinguish among source and target domain samples. The algorithm implements the following three phases (see Figure \ref{fig:ADDA}):
\begin{itemize}
\item \textit{Pre-training}: a source encoder \ac{CNN} together with a classifier are trained  in the classical way using the labeled source domain.
\item \textit{Adversarial Adaptation}: a target encoder \ac{CNN} is trained in such a way that a discriminator is not able to recognize the domain label of the examples.
\item \textit{Testing}: the classifier trained in the first phase is used together with the target mapping learned during the second phase to classify the target examples. 
\end{itemize}

\subsection{Source and target mapping}

The purpose of this method is to find a target mapping $M_t$ together with a classifier $C_t$ that is able to classify target images also if their labels are unknown (\textit{unsupervised adaptation}). The idea is to learn a source mapping $M_s$, a source classifier $C_s$ and then compute a good mapping $M_t$ for target domain adapting $M_s$ in such a way that target images can be efficiently classified using the source classifier $C_s$. The first choice to make is in the parametrization of these two mappings, keeping in mind that the distance between the source and target domains should be minimized while, at the same time, the category-discriminative target mapping should be preserved. The target mapping $M_t$ is implemented so that it matches, for the architecture, with the source mapping. Moreover, for each layer an equality constraint is imposed:
\begin{equation*}
\psi(M_d,M_t) \triangleq \{ \psi_{\ell_i}(M^{\ell_i}_s,M^{\ell_i}_t) \}_{i \in \{1...n\}}
\end{equation*}

where $M^{\ell_i}_s$ and $M^{\ell_i}_t$ are the $i$-th layer mapping parameters for the source and target domain respectively. The typical layerwise constraint that can be imposed in a \ac{CNN} with weight sharing is:

\begin{equation*}
\psi_{\ell_i} (M^{\ell_i}_s, M^{\ell_i}_t) = (M^{\ell_i}_s = M^{\ell_i}_t)
\end{equation*}

\subsection{Adversarial losses}

After that a parametrization for $M_t$ has been chosen, it is necessary to decide the adversarial loss for the target mapping. To train the generator in a \ac{GAN} is used the standard loss function but with reversed labels: 

\begin{equation*}
\mathcal{L}_{adv_M}(\textbf{X}_s,\textbf{X}_t,D) = - \mathbb{E}_{\textbf{x}_t \sim \textbf{X}_t}[\log D(M_t(\textbf{x}_t))]
\end{equation*}

We refer to this function with the expression \textit{\ac{GAN} loss function}. It can be noticed that with this formula only the target mapping $M_t$ is learned, the source mapping remains fixed. This is the typical approach of the \ac{GAN} method in which the distribution of real images is fixed and the distribution of the generate images is learned in order to match it.

\subsection{Adversarial discriminative domain adaptation}

We are able now to revisit the three phases of the \ac{ADDA} algorithm in a more technical way. The first phase, in which a source mapping $M_s$ and a source classifier $C$ ($C_s$) are learned, is performed in a standard way, with this classical supervised loss:

\begin{equation*}
\min_{M_s,C} \mathcal{L}_{cls}(\textbf{X}_s,Y_s) = - \mathbb{E}_{(x_s,y_s) \sim (\textbf{X}_s,Y_s) } \sum_{k=1}^K \mathbbm{1}_{[k=y_s]} \log C(M_s(\textbf{x}_s))
\end{equation*}

where $\mathbb{E}$ is the expected value of the source examples and $K$ is the number of class categories.

The second phase, in which a target mapping $M_t$ and a domain discriminator $D$ are learned, is performed alternately optimizing two loss functions:

\begin{equation}
\min_D \mathcal{L}_{adv_D} (\textbf{X}_s, \textbf{X}_t, M_s, M_t) = - \mathbb{E}_{\textbf{x}_s \sim \textbf{X}_s}[\log D(M_s(\textbf{x}_s))] - \mathbb{E}_{\textbf{x}_t \sim \textbf{X}_t}[\log (1 - D(M_t(\textbf{x}_t)))] 
\label{loss1}
\end{equation}

\begin{equation}
\min_{M_s,M_t} \mathcal{L}_{adv_M}(\textbf{X}_s,\textbf{X}_t,D) = - \mathbb{E}_{\textbf{x}_t \sim \textbf{X}_t}[\log D(M_t(\textbf{x}_t))]
\label{loss2}
\end{equation}

Since has been opted to let $M_s$ fixed during the second phase, the loss functions \eqref{loss1} and \eqref{loss2} are minimized without a revisiting of the pre-training phase.  

\chapter{Colorization methods}

\begin{figure}
\centering
\includegraphics[width=1\textwidth]{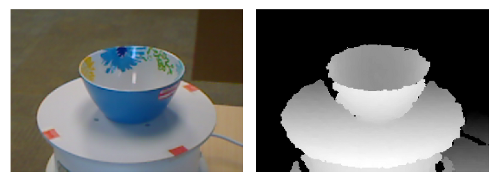}
\caption{\label{fig:rgbd} Example of RGB (left) and depth (right) data from the same object \cite{conf/icra/LaiBRF11}.}
\end{figure}

In this project the domain adaptation algorithms are applied using both RGB and depth informations as input data. It has been demonstrated that the depth produces additional informations to the standard RGB modality in the object recognition field: RGB provides texture, color and aspect informations while depth gives geometrical informations of the object shape that are invariant to the light conditions (see Figure \ref{fig:rgbd}). While RGB modality is composed by nature of three channels and it can be directly feed in a \ac{CNN}, depth informations needs some transformations to be mapped in the three-channel input of a network. This mapping is done to take advantage from the features of pre-trained \ac{CNN}s but also to match the two modalities and investigate the performance of the RGB-D input characteristics.

\section{Surface Normal Colorization}

\begin{figure}[h]
\centering
\includegraphics[width=0.4\textwidth]{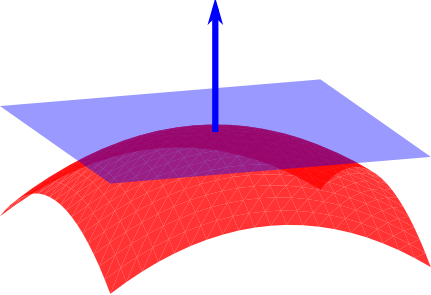}
\caption{\label{fig:normal} The normal of a point on a surface \cite{sn1}.}
\end{figure}

Normal maps are usually standard RGB images where the Red Green and Blue components coincide with the X, Y, and Z coordinates  of the surface normal. 
By definition a normal is a line or vector perpendicular to an object. In the three-dimensional space (our case of study) the normal is called \textit{surface normal} \cite{surfacenormal}. The surface normal of a point P belonging to an object is a vector perpendicular to the plane tangent to the surface in that point (see Figure \ref{fig:normal}).

\begin{figure}[h]
\centering
\includegraphics[width=0.35\textwidth]{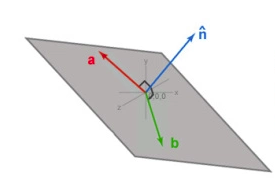}
\caption{\label{fig:norm} Surface normal geometric construction ($n=a \times b$) \cite{sn2}.}
\end{figure}

For each pixel of the depth image the surface normal is computed in this way: in the horizontal direction (x-axis) and in the vertical direction (y-axis) the gradients are computed to obtain two 3D vectors $a = [1,0, \frac{\partial z}{\partial x}]^T $ and $b = [0,1, \frac{\partial z}{\partial y}]^T $  towards the z-axis; the surface normal is computed with the cross product of $a$ and $b$ resulting in the vector $n = [-\frac{\partial z}{\partial x}, -\frac{\partial z}{\partial y},1]$ (see Figure \ref{fig:norm}). 

Once that the surface normal vector is computed, it is normalized using the Euclidean norm and then each of the three values of $n$ are mapped to the corresponding RGB channel in this way:

\begin{equation*}
x \in [-1,1] \rightarrow \text{\textbf{R}ed} \in [0,255]
\end{equation*}
\begin{equation*}
y \in [-1,1] \rightarrow \text{\textbf{G}reen} \in [0,255]
\end{equation*}
\begin{equation*}
z \in [0,1] \rightarrow \text{\textbf{B}lue} \in [128,255]
\end{equation*}

\begin{figure}[h]
\centering
\includegraphics[width=0.2\textwidth]{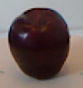}\quad\includegraphics[width=0.2\textwidth]{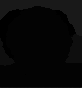}\quad\includegraphics[width=0.2\textwidth]{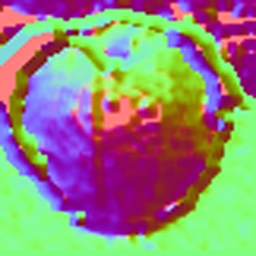}
\caption{\label{fig:examplesn} RGB image (left), original depth image (center), image of the depth mapped with Surface Normal (right).}
\end{figure}

An example of the final resulting depth image can be seen in Figure \ref{fig:examplesn}. We can notice that the surface normal colorization method capture properly the structural informations of the object.

\newpage
\section{Surface Normal$++$ Colorization}
To overcome the lack of informations that characterize most of the images of the datasets used in this project, for some experiment turned out to be useful to adopt a different pre-processing for the images called \textit{surface normal$\textit{++}$} \cite{aakerberg2017depth}.

The steps of image pre-processing are the following:

\begin{itemize}
\item \textit{Recursive median filter} proposed by (Lai et al., 2011). For each missing depth value of a depth image non-missing values in its neighbourhood are taken in consideration by a median filter that is recursively applied to fill the "holes" and to minimize the blurring. 

\item \textit{Borders underlined}. When a median filter is applied border problems may appear. To overcome this issue a border replication technique is used. 

\item \textit{Bilateral filter}. The task of this filter is to reduce the noise presents on a depth image preserving borders and increasing the smoothing.

\item \textit{Surface Normal}. For each pixel is applied the surface normal colorization explained in the previous section.

\item \textit{Unsharp mask filter}. When the bilateral filter is applied, despite trying to find a compromise between preserving borders and increasing the smoothing, some details are lost. To limit the damages an unsharp mask filter can be used. It increments contrast between borders and other high-frequency elements. 
\end{itemize}

\begin{figure}[h]
\centering
\includegraphics[width=1.1\textwidth]{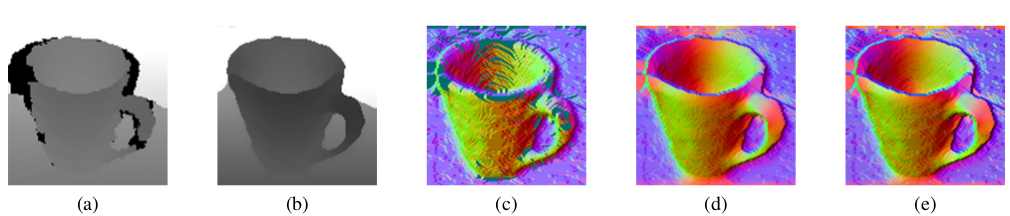}
\caption{\label{fig:normal_plus}Illustration of the pre-processing method steps. (a) Original depth image, (b) after applying recursive median filter, (c) after applying surface normal colorization, (d) application of surface normal colorization after the using of a bilateral filter, (e) after the application of an
unsharp mask filter \cite{aakerberg2017depth}.}
\end{figure}

In Figure \ref{fig:normal_plus} it can be seen an example of a depth image generated after the pre-processing described.

\chapter{Deep cue integration}

Due to the basic intuition under which a model trained with more features reaches a better performance, in the object recognition scenario the use of cue integration is becoming an increasingly popular choice. Particularly in this area characterized by a great intraclass variability, only one type of feature could not be enough to create a robust classifier that overcomes this issue: in many cases RGB information only could not be sufficient for the correct classification of an object whose shape, given by depth informations, could instead be decisive for the recognition.

\textbf{Classification with Multiple Cue Integration.}\newline
Let $\{ x_i,y_i\}_{i=1}^N$ be N samples of training set in which $x_i \in X$ is the input data (for instance an image) and $y_i \in Y$ is the label; let $\phi^j: X \rightarrow R^{D_j}$ be a features extractor from a set of $F$ functions ($j=1,\dots,F$) with $D_j$ the dimension of the j-th feature. The intention is to learn a classifier $f: X \rightarrow Y$ that is able to classify the new sample obtained by the integration of all the $F$ features extracted.

\section{Fusion Types}
\begin{figure}[h]
\centering
\includegraphics[width=0.7\textwidth]{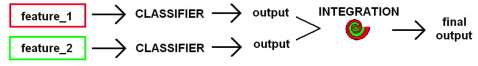}
\caption{\label{fig:highlevel}The high-level integration technique \cite{ci}.}
\end{figure}

In this section there will be a brief description of the three most popular techniques by which cue integration can be applied \cite{luo2011open}:

\begin{itemize}
\item \textbf{\textit{{Low-level Integration:}}} \newline 
the new samples are formed with the features extracted directly from data. It is also known as \textit{pre-mapping fusion} in fact the combination of data is done before any type of feature mapping using the information provided directly by sensors. Then, these new samples are put into a supervised learning algorithm to find the best parameters of the classifier (see Figure \ref{fig:lowlevel}). 

\begin{figure}[h]
\centering\includegraphics[width=0.6\textwidth]{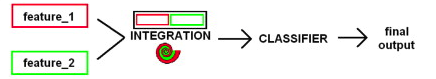}
\caption{\label{fig:lowlevel} The low-level integration technique \cite{ci}.}
\end{figure}

\item \textbf{\textit{{Middle-level Integration:}}} \newline
it is a more complex technique with respect to the other fusion approaches. In this case the integration is done in a middle level together with the mapping, then the resulted features are used for the final classifier (see Figure \ref{fig:mediumlevel}).

\begin{figure}[h]
\centering
\includegraphics[width=0.5\textwidth]{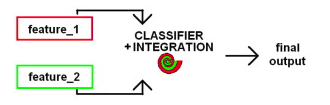}
\caption{\label{fig:mediumlevel}The middle-level integration technique \cite{ci}.}
\end{figure}

\item \textbf{\textit{{High-level Integration:}}} \newline
for each feature a classifier is trained independently, from each of them the confidence scores are extracted and combined to have the new samples. This type of integration can be seen as a two-layer scheme: with the first layer the confidence score for each feature is obtained using different learning algorithms; with the second layer these confidence scores are combined and used to train, for example, a linear \ac{SVM} (see Figure \ref{fig:highlevel}).
\end{itemize}

\newpage
\section{High-level Integration}

In the previous section a short overview has been made on the most popular approaches with which cue integration can be implemented. For the purpose of the project, now we focus on the last one: the \textit{high-level integration}.

Suppose that $\{ \textbf{x}_i,z_i \}_{i=1}^N$ is the training set, F is the number of cues chosen and $\phi^j$ is the j-th feature mapping. A classifier is trained for each input modality and then, to combine all the features together, for each modality the confidence scores $s^j(\textbf{x})$ associated to the specific classifier is taken.
Finally these confidence scores are put together for the final classification:

\begin{equation}
s(\textbf{x},z) = \sum_{j=1}^F \beta^j_z s^j(\textbf{x})
\label{eq}
\end{equation}

where $s^j(\textbf{x}) = \textbf{w}_z^j \cdot \phi^j(\textbf{x})$ with $\textbf{w}_z^j$ computed independently for each cue. The weights $\beta^j_z$ determine how much the j-th classifier should affect the final one, this quantity is optimized jointly considering all the classifiers involved (look at \cite{luo2011open} for more details). The high-level cue integration technique is introduced in this project with the aim of combine RGB and depth channels to evaluate the performance of this multi-modal data type in the domain adaptation scenario as in \cite{novi}. It is exploited following this sequence: 

\begin{itemize}
\item One network, properly modified to perform domain adaptation, is trained independently for RGB and depth input modalities in such a way that the best weights $\textbf{w}_z^j$ (with $j=1,2$) are found for both data type.
\item The activation values (that represent the confidence score $s^j(\textbf{x})$ in \eqref{eq}) of the last layer before the classification (for example the output of the layer fc7 in \textit{AlexNet}) are taken from the two trained network. For each image, of both source and target domain, two feature vectors (one for RGB modality, one for depth modality) are captured. Subsequently they are concatenated having in the end one "big" feature vector for each image (that intrinsically has both RGB and depth informations). 
\item The new concatenated features of source domain are used to train the linear \ac{SVM} and the new concatenated features of target domain are used to test it for evaluate the adaptation.
\end{itemize}

\newpage
\chapter{Experiments}
\section{Datasets}

All the unsupervised adaptation transfer tasks are conducted on five publicly-available datasets: \ac{ROD}, \ac{ARID}, \ac{WOD}, \ac{BigBIRD} and Active Vision Dataset.

\subsection{RGB-D Object Dataset}
\begin{figure}[h]
\centering
\includegraphics[width=0.8\textwidth]{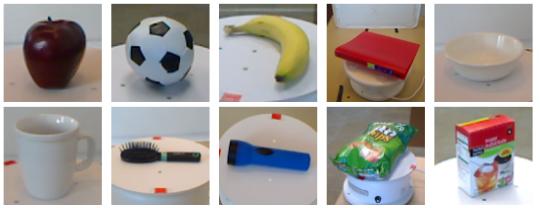}
\caption{\label{fig:washington}Objects from the RGB-D Object Dataset \cite{conf/icra/LaiBRF11}.}
\end{figure}

\ac{ROD} is composed by objects of everyday life that can be found in places where robots should be able to operate (home, office, \dots) (see Figure \ref{fig:washington} for some examples).

\begin{figure}[h]
\centering
\includegraphics[width=0.8\textwidth]{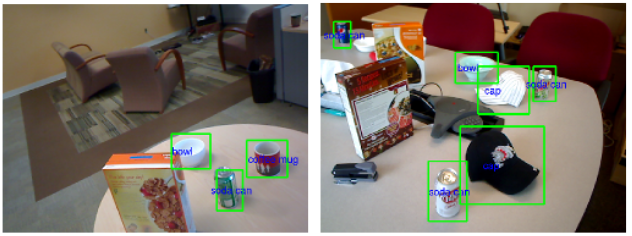}
\caption{\label{fig:wash2}Detection results in two multi-object scenes \cite{conf/icra/LaiBRF11}.}
\end{figure}

The categories of objects involved are 51, a subset of the 1000 in ImageNet \cite{Deng09imagenet:a}. In particular, the dataset has visual and depth informations about 300 objects each belonging to one of these 51 categories (as result one category has from three to fourteen instances). The images of the objects have been captured by an RGB-D camera that together get color and depth informations at 640x480 pixels of resolution; each pixel has actually four channels: red, blue, green and depth. The camera has been located one meter away from a turntable that rotates with a fixed speed on itself. Registration took place with the camera mounted at 30°, 45° and 60° above the horizon; for each object, data have been recorded at 20 Hz during a complete revolution at each of the three heights. The dataset also provides 8 natural scenes of ordinary indoor places in which the objects of the RGB-D dataset have been disseminated, the scenes were recorded using the RGB-D camera at an height almost equal to the level of the eyes of a human being. Together with these 8 natural scene, also the ground truth bounding boxes are provided (see Figure \ref{fig:wash2}).

Extracting the images from these scene you will have images of the objects in different ways: from different points of view, at a different distance from the camera or with partial occlusion.

\subsection{Autonomous Robot Indoor Dataset}
\begin{figure}[h]
\centering
\includegraphics[width=0.7\textwidth]{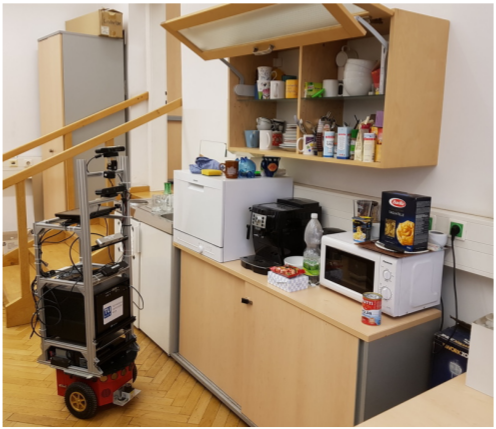}
\caption{\label{fig:arid} The mobile robot that acquire data from a scene with everyday objects \cite{Arid}.}
\end{figure}

\ac{ARID} coincides in number and typology with the 51 categories of \ac{ROD} described in the previous paragraph. This two datasets can be seen as complementary: the same objects are collected in a restricted setting (in this case we refer only to the raw \ac{ROD} without the natural scenes) and in a real-life environment (\ac{ARID}). The real-life context, with which a robot should be able to deal, is obtained acquiring the images with an RGB-D camera attached on a mobile robot that navigates in a typical human environment (see Figure \ref{fig:arid}).

Hence, despite \ac{ROD} that is characterized by an assigned camera-object distance, a fixed background and no changes in light setting, \ac{ARID} includes all these real-world traits: alteration in the objects illumination, dynamic viewpoint, clutter, occlusion, partial view, change in the object scale and background variation. The total number of objects involved is 153: 3 instances for each category. To guarantee the natural variation in the lighting of the objects, data have been acquired in 10 sessions not in the same day and in different hours. Each session has the duration of one hour in which the mobile robot loops over four established waypoints. The environment has been prepared with 30-31 objects that are relocated every two patrolling loops to assure the variation of both the object view and the camera-object distance. Every time that the mobile robot reaches a waypoint the RGB-D camera scans the scene using the pan-tilt unit to do an horizontal movement. The camera collects RGB and depth images at 30 Hz with a resolution of 640x480 pixels.

\subsection{Web Object Dataset}
\ac{WOD} is composed by images taken from the web through searches query made on Google, Yahoo, Bing and Flickr. The best images among those resulting from the researches are selected using a method designed by Massouh et al. \cite{massouh17} that use both visual and \ac{NLP} informations to eliminate most of the noise (the remainder is removed by hand). The objects that constitute the dataset belong to the same 51 categories of \ac{ROD} and \ac{ARID} but, unlike them, \ac{WOD} has a much bigger number of instances (each image potentially contains a different object) and he does not have depth informations of the objects.

\subsection{Big Berkeley Instance Recognition Dataset}
\begin{figure}[h]
\centering
\includegraphics[width=0.4\textwidth]{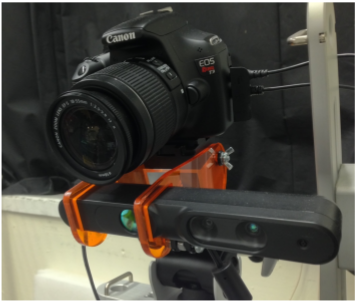}
\caption{\label{fig:bigbird3}RGBDToolkit \cite{BigBird}}
\end{figure}

\begin{figure}[h]
\centering
\includegraphics[width=0.6\textwidth]{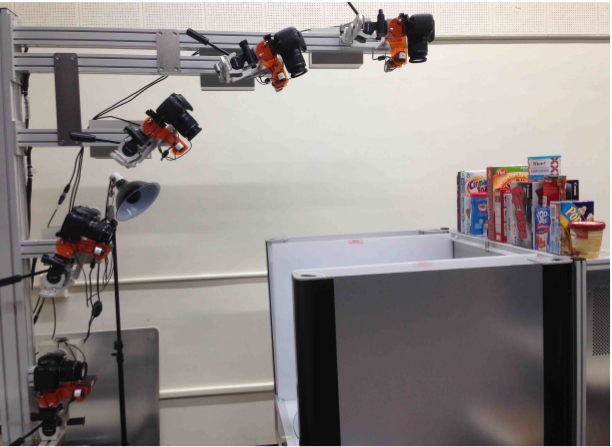}
\caption{\label{fig:bigbird}Side view of the five cameras and the five depth sensors mounted ready to start a data acquisition process \cite{BigBird}.}
\end{figure}

\ac{BigBIRD} was designed with the aim of investigate object recognition only at instance level. It is composed by 100 3D object instances each different from the others but without any type of classification at category level. It's an high quality dataset both for the large number of images per object (with RGB and depth informations) and for the high-resolution that characterizes them. The images were captured using 5 high-resolution cameras (12.2 MP) and 5 depth sensors mounted using a RGBDToolkit (see Figure \ref{fig:bigbird3}) in 5 different heights and positions (see Figure \ref{fig:bigbird}). 

\begin{figure}[h]
\centering
\includegraphics[width=0.6\textwidth]{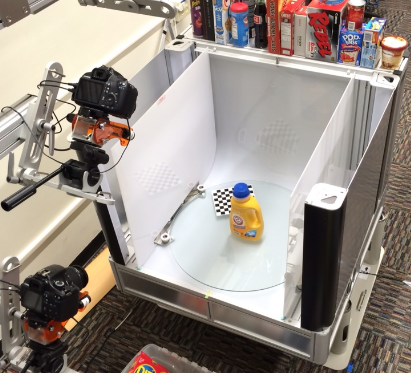}
\caption{\label{fig:bigbird2}The glass turntable of the data-collection system \cite{BigBird}.}
\end{figure}

Each object has been placed on the glass turntable to which the cameras, together with four lights, points (see Figure \ref{fig:bigbird2}).

The four lights are located at the bottom, at the back wall, at the front corners and at the back corners. It is also important to point out the presence of a chessboard that improves data calibration between RGB and depth informations. One by one the objects have been placed on the turntable which rotated in units of 3 degree until done a complete revolution and, for each angle, all five cameras acquire data. In total 600 RGB images and 600 point clouds have been acquired for each object ($\ang{360}/\ang{3}$ = 120, 120 $\times$ 5 = 600). Subsequently a segmentation mask of the objects, for each view, was produced. The entire process of acquisition of the data took less than 6 minutes for object almost without human contribution.

\subsection{Active Vision Dataset}
\begin{figure}[h]
\centering
\includegraphics[width=0.9\textwidth]{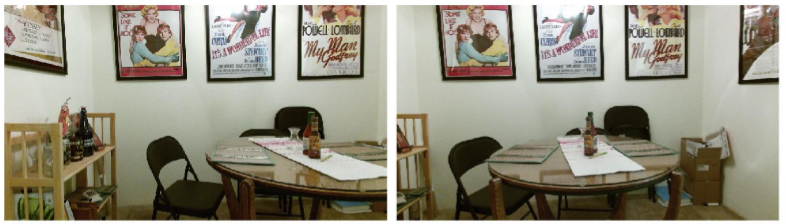}
\caption{\label{fig:activevision2} Example of scenes of the dataset \cite{ActiveVision}}
\end{figure}

Active Vision Dataset was built starting from the \ac{BigBIRD} dataset described above: 33 object instances similar to those of \ac{BigBIRD} are included in the scenes. The number of scenes recorded is 9 (see Figure \ref{fig:activevision2} for examples) but some of them are captured twice with some small changes (displacing objects that are typically moved by people as books, chairs, \dots) for a total of 17 scans. 

\begin{figure}[h]
\centering
\includegraphics[width=0.6\textwidth]{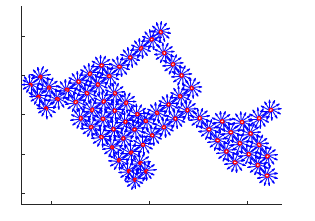}
\caption{\label{fig:activevision}Map of the movements of the camera in a scene with all the subset of point chosen for the data recording \cite{ActiveVision}.}
\end{figure}

The goal is to resemble the motion of a robot that moves within daily environments such as office, kitchen, living room, ecc.. The scenes are recorded within these type of rooms with a Kinect v2. For each scene a set of points (58-201) is chosen, once the robot has reached one of these points, it will rotate the camera on itself. It was chosen to record an image every 30 degree in each point to not have an excessive number of images for scene (see Figure  \ref{fig:activevision}). The scans have been labeled resulting in 3000 (on average) 2D bounding box for each scan.

\newpage
\section{Experimental Setup}

The experiments were conducted using the four deep domain adaptation networks described in chapter 3: \ac{DAN} \cite{DAN}, \ac{DANN} \cite{DANN}, \ac{AutoDIAL} \cite{autoDIAL}, \ac{ADDA} \cite{ADDA}. The evaluation is performed using multimodal input data: RGB only, depth only and RGB-D on the adaptation tasks \hbox{\ac{ROD} $\rightarrow$ \ac{ARID}},  \hbox{\ac{WOD} $\rightarrow$ \ac{ARID}} and  \hbox{\ac{BigBIRD} $\rightarrow$ Active Vision}. The domain adaptation algorithms have been implemented modifying properly two of the most popular deep network architectures: AlexNet \cite{krizhevsky2012imagenet} and ResNet-50 \cite{he2016deep}. In both cases the networks are pre-trained on ImageNet \cite{deng2009imagenet} and then fine-tuned on the specific dataset. Depth images have been colorized either with Surface Normal, in the case of depth only input modality, or with Surface Normal$++$, for the RGB-D data (both colorization methods are described in chapter 4); the resulted images, now mapped on three channels, can be fed into domain adaptation networks as if they were RGB images.\newline 

\begin{figure}[h]
\centering
\includegraphics[width=0.31\textwidth]{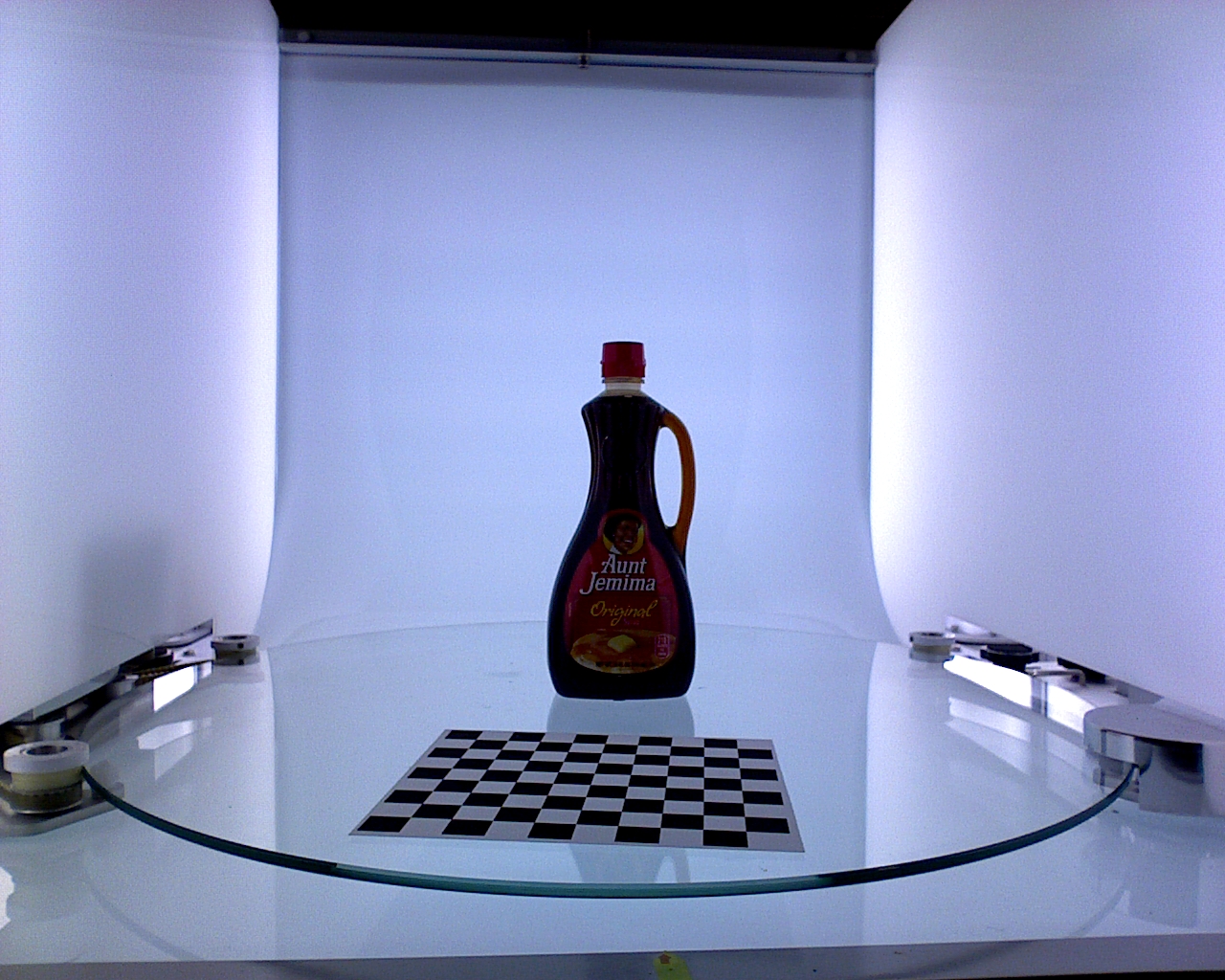}
\quad\includegraphics[width=0.31\textwidth]{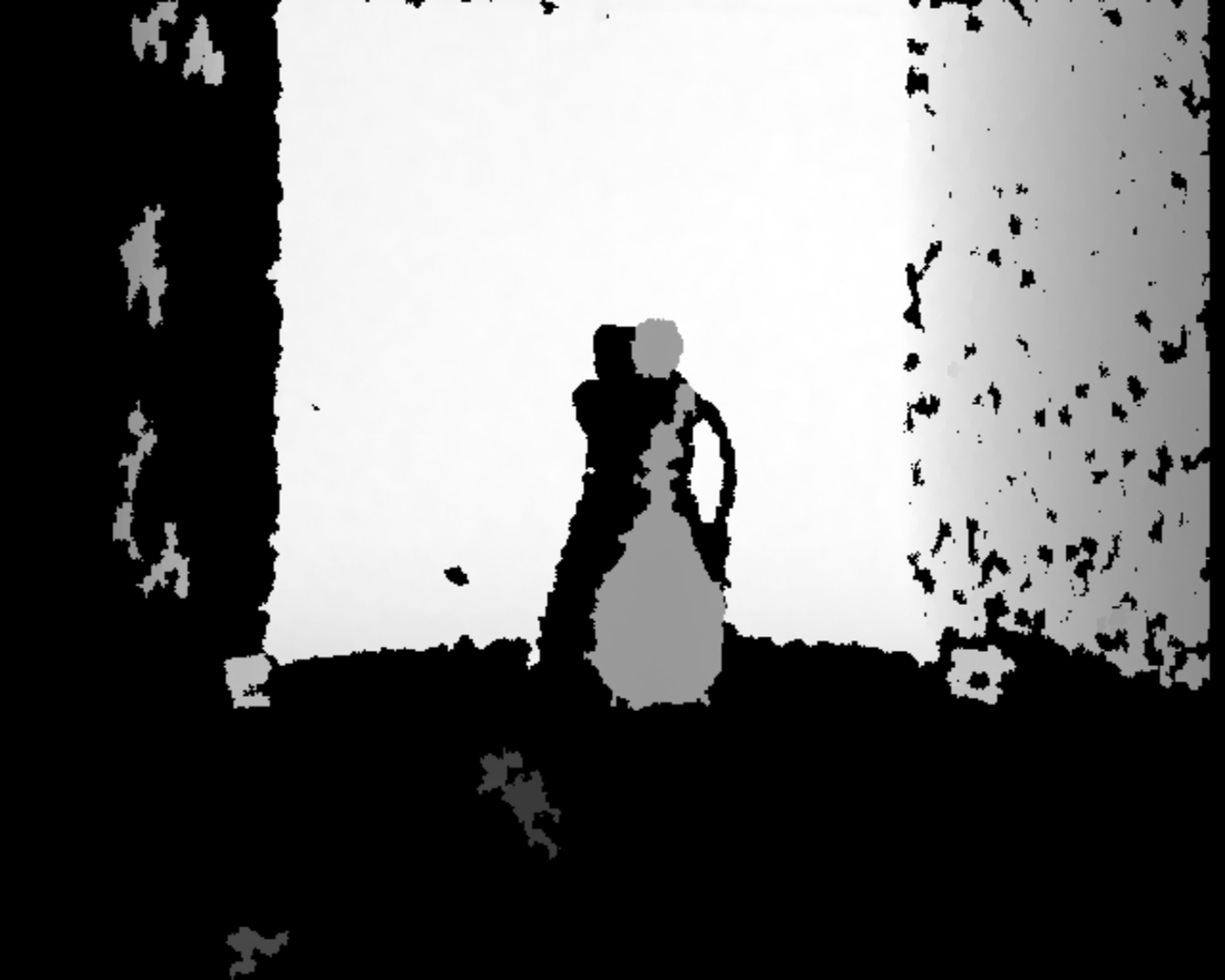}\quad\includegraphics[width=0.31\textwidth]{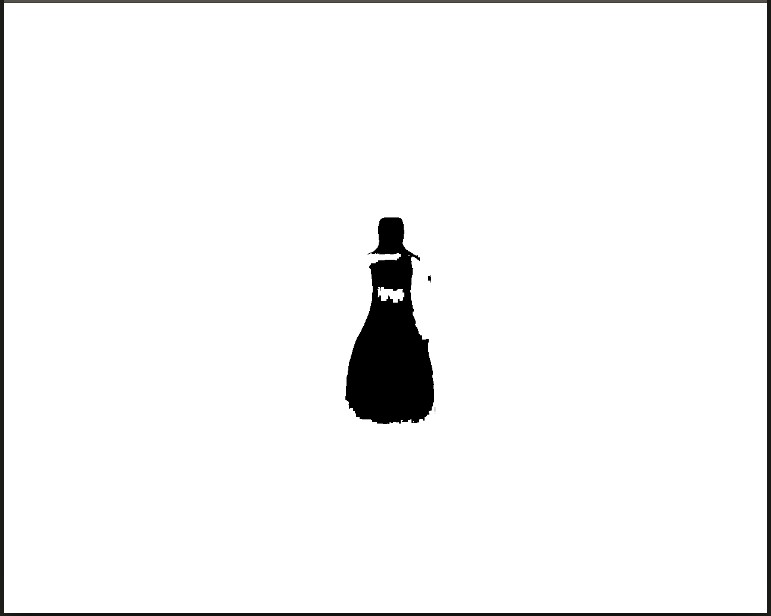}\caption{\label{fig:mask}From left to right: RGB image, depth image, segmentation mask provided by Big Berkeley Instance Recognition Dataset.}
\end{figure}

\textbf{Datasets}. In order to have datasets belonging to the same adaptation task approximately of the same size, from \ac{ROD} 41877 images have been taken in RGB and depth formats, from \ac{ARID} 40713 and from \ac{WOD} 50547 (for this last dataset only in RGB format, the reason is explained in the previous section). It is noteworthy that for the RGB-D experiments, to reach a better performance, only a subset of \ac{ARID} composed by 36050 images has been used: depth images with more than $75\%$ of null pixels have been removed. For \ac{BigBIRD} and Active Vision, with the aim of have the same amount of images for both datasets with a balanced distribution over all the object classes, a particular selection was done. First of all the evaluation set of \ac{BigBIRD}, composed by only a subset of the whole dataset, has been selected with the aim of avoiding the presence of too similar images that could compromise the training with overfitting. Applying the same selection policy used for the evaluation set of Washington, for each object one frame every 6 degree of rotation, from each of the 5 cameras, is taken obtaining a total of 300 images per class ($\ang{360}$/\ang{6} = 60, $60 \times 5=300$). Then, since the images have a lot of noise (the chessboard, the glass turntable, etc.) it has been decided to make a crop focused on the object. From the official website you can download the RGB images, the depth images and the segmentation masks (an example in Figure \ref{fig:mask}), the crop is done using the segmentation mask to infer border informations about the object.

\begin{figure}[h]
\centering
\includegraphics[width=0.15\textwidth]{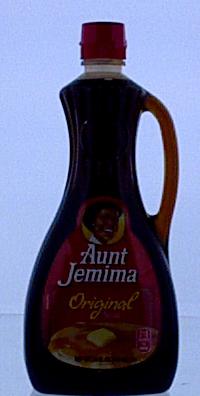}
\quad\includegraphics[width=0.15\textwidth]{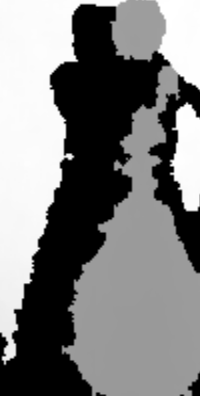}
\caption{\label{fig:depth_cropped}Example of a cropped RGB image on left, example of a cropped depth image on right.}
\end{figure}

For all the images (in both RGB and depth formats) it was decided to have a bounding box that wrap the object in such a way that there are 25 pixel from the borders of it to the edges of the box (see Figure \ref{fig:depth_cropped} for an example of both formats). 

Since some objects appear very rarely in Active Vision scenes, to have a dataset composed by a sufficient quantity of images for each object class (300, to be consistent with \ac{BigBIRD}), the number of categories in these two datasets has been reduced from 33 to 30. The objects removed are: \textit{expo marker red} (221 images), \textit{softsoap clear} (286 images) and \textit{red cup} (12 images). For Active Vision the 300 images were selected randomly among all those available in each category. After these steps both datasets become composed by 9000 images in either RGB and depth formats.\newline

\textbf{Preprocessing.} It was done a different preprocessing of the images in function of the starting network used: for AlexNet both RGB and depth images are first scaled to 256x256 pixels and then randomly cropped at 227x227 pixels; for ResNet-50 it was necessary to do the preprocessing of RGB and depth images in a specular way without any random component since the weights computed using this basic network are then used also for RGB-D experiments. In this case the images are scaled directly to 224x224 pixels that is the size of the input data layer of ResNet-50.\newline

\textbf{DAN setting}. For AlexNet the \ac{MK-MMD} adaptation regularizer is added to the \ac{CNN} risk for \textit{fc7-fc8} layers; in ResNet-50 it is added for \textit{average pool} and \textit{fc1000}. For all the experiments the parameters related to the learning rate policy have been set up as proposed by the authors: policy \textit{inverse} \footnote{

\begin{equation*}
\text{learning\_rate} = \text{base\_lr} \cdot (1+\gamma \cdot \text{current\_iteration}) ^ {- \text{power}}
\end{equation*}

}, with gamma 0.001, power 0.75 and the back propagation method used to train the network is \ac{SGD} with momentum 0.9.\newline

\textbf{DANN setting}. The \ac{GRL} is added after \textit{fc7} in AlexNet and after \textit{pool5} in ResNet-50. Also for \ac{DANN} the learning rate policy parameters have been set up as proposed by the authors: policy \textit{inverse}, with gamma 0.001, power 0.75 and the back propagation method used to train the network is \ac{SGD} with momentum 0.9.\newline

\textbf{AutoDIAL setting.} The Domain Alignment layers are 
inserted in \hbox{\textit{fc6-fc7-fc8}} for AlexNet and only in \textit{fc1000} for ResNet-50. The learning rate policy has been set up also in this case as the authors: policy \textit{step} \footnote{\begin{equation*}
\text{learning\_rate} = \text{base\_lr} \cdot \gamma^{\frac{\text{current\_iteration}}{\text{size\_step}}}
\end{equation*}}, gamma 0.1, step size $85\%$ of the total number of epochs, weight decay $5 \times 10^{-4}$ and the back propagation method used to train the network is \ac{SGD} with momentum 0.9.\newline

\textbf{ADDA setting.} In this case only ResNet-50 has been used as base model, the algorithm is applied on this network with the same setting proposed by the authors. In particular, for the second phase of the algorithm (the \textit{Adversarial Adaptation}), the layers up to block \textit{conv4} of the target model have been fine-tuned. The \textit{discriminator} is composed by three fully connected layer of size 1024, 2048 and 3072 every succeed by ReLUs and there is one fully connected layer for the final outcome. The policy of the learning rate is \textit{fixed}\footnote{\begin{equation*}
\text{learning\_rate} = \text{base\_lr} 
\end{equation*}} and the back propagation method used to train the network is \ac{SGD} with momentum 0.9.\newline

\textbf{Training}. To have a starting point for the evaluation of the results obtained from the domain adaptation tasks a baseline for each input modality is also provided. It consists of the results obtained training a network with source domain and testing it on target domain without any kind of adaptation. The set of adaptation experiments can be divided into two macro groups: (1) whole target dataset in both train and test phases, (2) two different subsets of target dataset for train and test phases. For group (1) the size of target datasets have been shown in the paragraph \textit{Datasets} of 
this section: \hbox{40713 $\rightarrow$ \ac{ARID}}, 
\hbox{9000 $\rightarrow$ Active Vision}. For group (2) a subdivision of them was made. Since \ac{ARID} is provided with 3 instances for each object class, the dataset has been divided into three different train/test splits using for each test set a different instance of the same object. The size of the resulted splits obviously is variable: Split1 $\rightarrow$ 26188/14525, Split2 $\rightarrow$ 27685/13028, Split3 $\rightarrow$  27553/13160. Active Vision is subdivided according to another policy but also in this case the splits created are three. Each split has the same size: 7200 images for train and 1800 images for test. The subdivision is done, for the test set, taking 60 images for
each object class \hbox{(60 $\times$ 30 = 1800)} and, for the train set, the 
rest of the images have been taken \hbox{( $(300-60) \times 30 = 7200$ )}. Obviously 
the images chosen for the test sets are different for each split.

\newpage
\section{Results and Discussion}
The experiments have been conducted trying several base learning rates, different batch sizes and seldom also trying different configurations of the learning rate multipliers within the networks. Below are shown only the parameter settings related to the best results obtained. For all the experiments the mean file of ImageNet dataset has been used.

\subsection{RGB Only}
Tables \ref{tab:rgbgroup1} and \ref{tab:rgbgroup2} show the best results for the adaptation tasks \hbox{\ac{ROD}/\ac{WOD} $\rightarrow$ \ac{ARID}} relatively for the group of experiment (1) and (2) (the subdivision of the experiments is explained in the previous section in paragraph \textit{Training} of Section 6.2) as well, Tables \ref{tab:rgbgroup1bb} and \ref{tab:rgbgroup2bb} report the best results for the adaptation task \hbox{\ac{BigBIRD} $\rightarrow$ Active Vision}. The parameter setting which led to the best results for the shift \uline{\hbox{\ac{ROD}/\ac{WOD} $\rightarrow$ \ac{ARID}}} in both groups of experiments is the following:
\begin{itemize}
\item \uline{\ac{DAN}} \newline \hbox{AlexNet $\rightarrow$ lr\tablefootnote{base learning rate}: 0.0001, bs\tablefootnote{batch size}: 64, epochs: 30, \ac{MK-MMD} loss weight: 1} \newline
\hbox{ResNet-50 $\rightarrow$ lr: 0.001, bs: 128, epochs: 30, \ac{MK-MMD} loss weight: 0.3}

\item \uline{\ac{DANN}}\newline
\hbox{AlexNet $\rightarrow$ lr: 0.0001, bs: 128, epochs: 30, Domain Classifier loss weight: 0.1} \newline
\hbox{ResNet-50 $\rightarrow$ lr: 0.001, bs: 64, epochs: 30, Domain Classifier loss weight: 0.1}

\item \uline{\ac{AutoDIAL}}\newline
AlexNet $\rightarrow$ lr: 0.0001, bs: 256, epochs: 30, Target loss weight ($\lambda$ in \eqref{lossautodial}): 0.1 \newline
ResNet-50 $\rightarrow$ lr: 0.001, bs: 256, epochs: 30, Target loss weight ($\lambda$ in \eqref{lossautodial}): 0.4

\item \uline{\ac{ADDA}}\newline
ResNet-50 $\rightarrow$ lr: 0.001, bs: 128, epochs: 30, Mapping \& Adversarial loss weight: 1

\end{itemize}
We can notice that, with AlexNet as basic network, \ac{AutoDIAL} outperforms all comparison methods; using ResNet-50, instead, the best algorithm is \ac{DANN} although also the other algorithms obtained good results. Contrary to expectations, only the half of the second group of experiments has a worse performance with respect to the first group although in the second group the adaptation and the test phases have been done with a different set of images. In this second group of experiments it can be notice that the best performance using AlexNet is obtained with \ac{DANN}, instead, using ResNet-50 \ac{AutoDIAL} reaches the best accuracy. It is interesting to note that the highest level of accuracy, for ResNet-50, with the second group of experiments is reached. It was chosen to perform the adaptation for \hbox{\ac{WOD} $\rightarrow$ \ac{ARID}} using only the algorithm which led to the best result in the shift \hbox{\ac{ROD} $\rightarrow$ \ac{ARID}}: \ac{DANN}. We can notice an evident improvement of the accuracy obtained using \ac{WOD} as source dataset. This is motivated by the fact that taking the images from the web gives rise to a much more realistic dataset with respect to \ac{ROD} in which the images are acquired in a quite unrealistic settings. 

\begin{table}
\centering
\begin{tabular}{lccc}
&\multicolumn{1}{c}{\textbf{AlexNet}}&\multicolumn{2}{c}{\textbf{ResnNet-50}}\\
Method & ROD $\rightarrow$ ARID & ROD $\rightarrow$ ARID & WOD $\rightarrow$ ARID\\ \hline\hline
Source only \cite{Arid} & 0.291 & 0.337 &0.388 \\
DAN &  0.34 &  0.429 & -\\
DANN &  0.329 &\textbf{0.459} &\textbf{0.582}\\
AutoDIAL & \textbf{0.378} & 0.442 & -\\
ADDA & - & 0.422 & -
\end{tabular}
\caption{\label{tab:rgbgroup1} ROD/WOD $\rightarrow$ ARID RGB experiments of group (1).}\quad \begin{tabular}{lccc}
&\multicolumn{1}{c}{\textbf{AlexNet}}&\multicolumn{2}{c}{\textbf{ResnNet-50}}\\
Method & ROD $\rightarrow$ ARID & ROD $\rightarrow$ ARID & WOD $\rightarrow$ ARID\\ \hline\hline
Source only \cite{Arid} & 0.291 & 0.337 &0.388 \\
DAN &  0.349 &  0.466  & -\\
DANN &  \textbf{0.356} &0.439 &\textbf{0.527}\\
AutoDIAL & 0.299 & \textbf{0.467} & -\\
ADDA & - & 0.357 & -
\end{tabular}
\caption{\label{tab:rgbgroup2}ROD/WOD $\rightarrow$ ARID RGB experiments of group (2).}
\end{table}

For the shift \uline{\hbox{\ac{BigBIRD} $\rightarrow$ Active Vision}} the best set of parameters is the following:
\begin{itemize}
\item \uline{\ac{DAN}} \newline \hbox{AlexNet $\rightarrow$ lr: 0.001, bs: 128, epochs: 30, \ac{MK-MMD} loss weight: 0.7} \newline
\hbox{ResNet-50 $\rightarrow$ lr: 0.001, bs: 64, epochs: 30, \ac{MK-MMD} loss weight: 0.7}

\item \uline{\ac{DANN}}\newline
\hbox{AlexNet $\rightarrow$ lr: 0.001, bs: 64, epochs: 30, Domain Classifier loss weight: 0.1} \newline
ResNet-50 $\rightarrow$ lr: 0.001, bs: 128, epochs: 30, Domain Classifier loss weight: 0.1

\item \uline{\ac{AutoDIAL}}\newline
AlexNet $\rightarrow$ lr: 0.001, bs: 256, epochs: 30, Target loss weight ($\lambda$ in \eqref{lossautodial}): 0.1 \newline
ResNet-50 $\rightarrow$ lr: 0.001, bs: 256, epochs: 30, Target loss weight ($\lambda$ in \eqref{lossautodial}): 0.1

\item \uline{\ac{ADDA}}\newline
ResNet-50 $\rightarrow$ lr: 0.001, bs: 128, epochs: 30, Mapping \& Adversarial loss weight: 1
\end{itemize}

We can observe that, for the first group of experiments, the best algorithms, for both AlexNet and ResNet-50 networks, are the same as for the task \hbox{\ac{ROD} $\rightarrow$ \ac{ARID}}, but here \ac{AutoDIAL} and \ac{DANN} remain the best algorithms also for the second group of experiments. As for the previous adaptation shift, paradoxically, the performances increase in the second group of experiments for almost all the algorithms. Probably this is due to the fact that, using less images from the target domain during the adaptation, a more general classifier is produced.

\begin{table}
\centering
\begin{tabular}{lcc}
&\multicolumn{1}{c}{\textbf{AlexNet}}&\multicolumn{1}{c}{\textbf{ResnNet-50}}\\
Method & BigBIRD $\rightarrow$ Active Vision & BigBIRD $\rightarrow$ Active Vision\\ \hline\hline
Source only & 0.264 & 0.361\\
DAN &  0.459 &  0.497 \\
DANN &  0.473 & \textbf{0.537}\\
AutoDIAL & \textbf{0.521} & 0.512 \\
ADDA & - & 0.512
\end{tabular}
\caption{\label{tab:rgbgroup1bb}BigBIRD $\rightarrow$ Active Vision RGB experiments of group (1).}\quad\begin{tabular}{lcc}
&\multicolumn{1}{c}{\textbf{AlexNet}}&\multicolumn{1}{c}{\textbf{ResnNet-50}}\\
Method & BigBIRD $\rightarrow$ Active Vision & BigBIRD $\rightarrow$ Active Vision\\ \hline\hline
Source only & 0.264 & 0.361\\
DAN &  0.344 &  0.518 \\
DANN &  0.499 & \textbf{0.542}\\
AutoDIAL & \textbf{0.558} & 0.498 \\
ADDA & - & 0.572
\end{tabular}
\caption{\label{tab:rgbgroup2bb}BigBIRD $\rightarrow$ Active Vision RGB experiments of group (2).}
\end{table}

\newpage
\subsection{Depth Only}

\begin{figure}[h]
\centering
\includegraphics[width=0.9\textwidth]{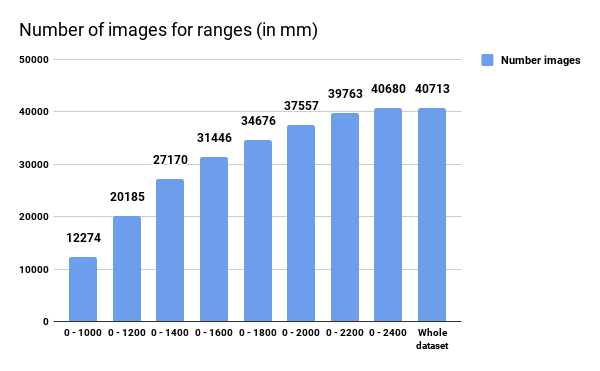}
\caption{\label{fig:all_number_images}Quantity of images for each distance range in Autonomous Robot Indoor Dataset dataset.}
\end{figure}

\begin{figure}
\centering
\includegraphics[width=0.9\textwidth]{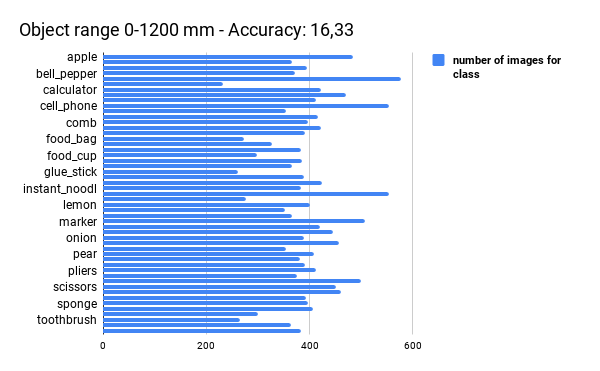}\quad
\includegraphics[width=0.9\textwidth]{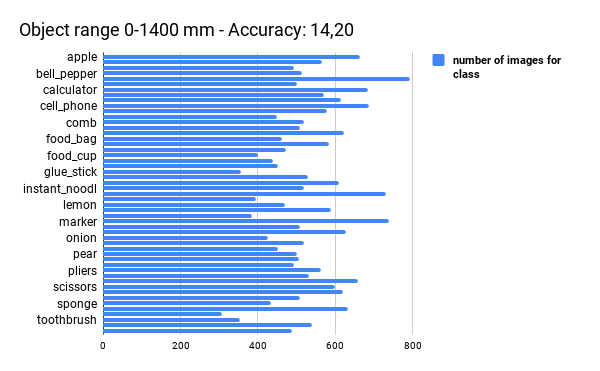}\quad
\caption{\label{fig:class_distribution1} Quantity of images for each object class in ranges: 0-1200/0-1400.}
\end{figure}

\begin{figure}
\centering
\includegraphics[width=0.9\textwidth]{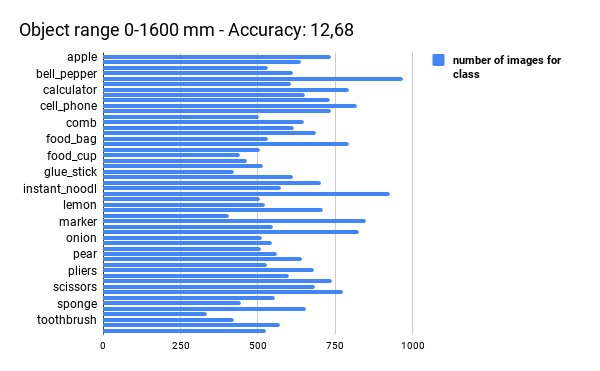}\quad
\includegraphics[width=0.9\textwidth]{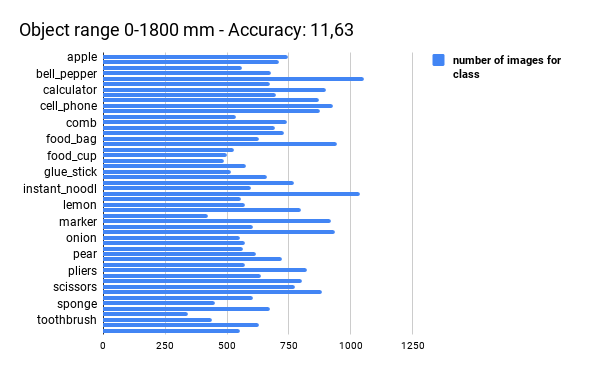}
\caption{\label{fig:class_distribution2} Quantity of images for each object class in ranges: 0-1600/0-1800.}
\end{figure}

\begin{figure}
\centering
\includegraphics[width=0.9\textwidth]{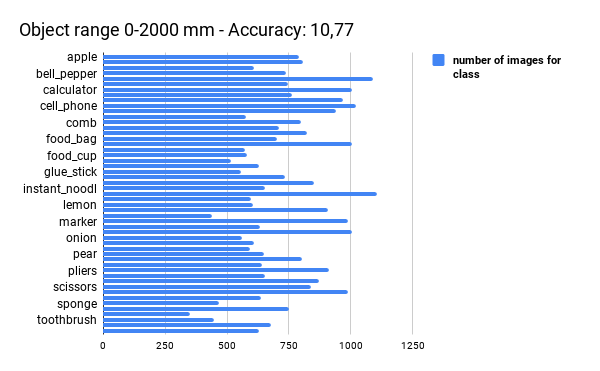}\quad
\includegraphics[width=0.9\textwidth]{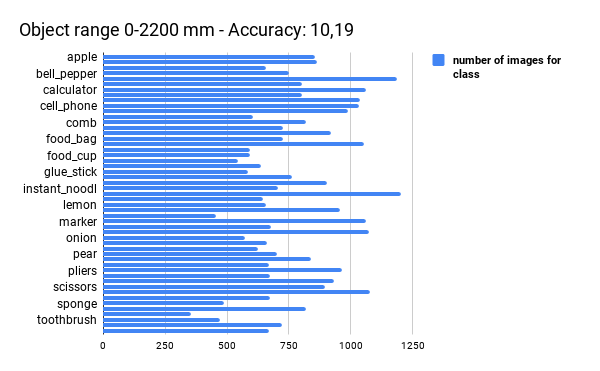}
\caption{\label{fig:class_distribution3} Quantity of images for each object class in ranges: 0-2000/0-2200.}
\end{figure}

\begin{figure}
\centering
\includegraphics[width=0.9\textwidth]{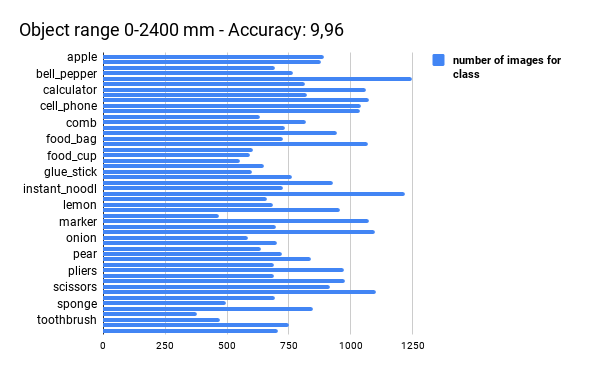}\quad
\includegraphics[width=0.9\textwidth]{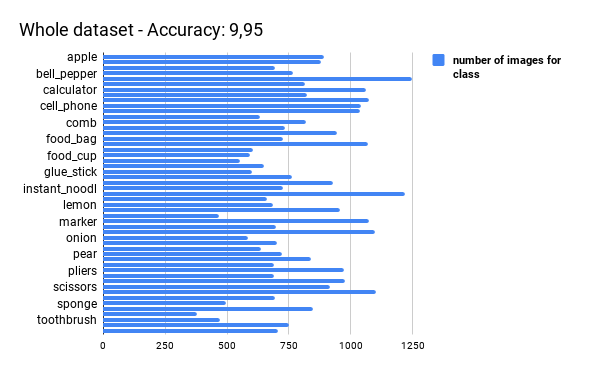}
\caption{\label{fig:class_distribution4} Quantity of images for each object class in ranges: 0-2400/0-max distance.}
\end{figure}

As said in the previous section, a baseline is obtained testing the target domain on a network trained with source domain in absence of adaptation. About the baseline for the task \hbox{\ac{ROD} $\rightarrow$ \ac{ARID}} it is interesting to do experiments based on the distance of the objects from the camera for two reasons: (i) RGB-D cameras have specific ranges of distance within which they provide reliable results, (ii) \ac{ROD} dataset contains only images with objects in about a meter from the camera so, the classifier trained on this dataset should be able to recognize better the images of \ac{ARID} in which the objects are within one meter from the observer.
With this in mind, eight subsets of images from \ac{ARID} have been grouped. 
The ranges considered are: \hbox{0 - 1200 mm}, \hbox{0 - 1400 mm}, \hbox{0 - 1600 mm}, \hbox{0 - 1800 mm}, \hbox{0 - 2000 mm}, \hbox{0 - 2200 mm}, \hbox{0 - 2400 mm}, whole dataset. Figure \ref{fig:all_number_images} shows the histogram of the quantity of images for each range of subdivision and, for completeness, Figure \ref{fig:class_distribution1}, \ref{fig:class_distribution2}, \ref{fig:class_distribution3}, \ref{fig:class_distribution4} report the distribution of the objects in function of their category for each specific range. 

\begin{figure}[h]
\centering
\includegraphics[width=0.9\textwidth]{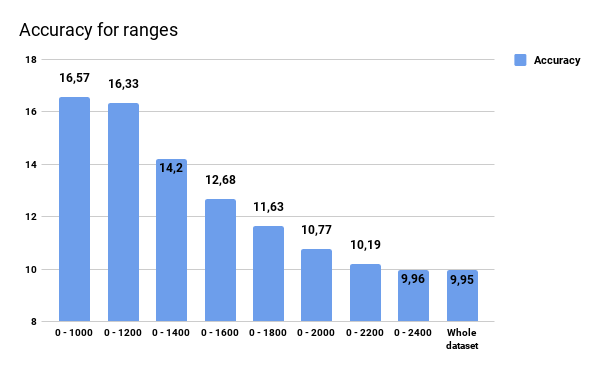}
\caption{\label{fig:all_accuracies}Accuracies obtained testing the network trained on RGB-D Object Dataset on the eight subsets of Autonomous Robot Indoor Dataset.}
\end{figure}

Testing a network (AlexNet), trained with \ac{ROD}, on each of the eight subsets, the accuracy decreases with the increase of distance range (see Figure \ref{fig:all_accuracies}). As predicted, the best accuracy is obtained using as test dataset the images of \ac{ARID} in which the images are within a meter of distance from the camera. This is a limitation of \ac{ROD} dataset due to a too static recording. Although it is not the best, to be consistent with RGB experiments, as baseline is taken the result obtained with the  whole \ac{ARID} dataset.

Now we are ready to discuss the results obtained applying domain adaptation algorithms to colorized depth input data. Table \ref{tab:depth1} and \ref{tab:depth2} show the unsupervised adaptation results of the transfer task \uline{\hbox{\ac{ROD} $\rightarrow$ \ac{ARID}}} for both (1) and (2) groups of experiments. The parameter setting which led to these results, in both groups of experiments, is the following:
\begin{itemize}
\item \uline{DAN} \newline \hbox{AlexNet $\rightarrow$ lr: 0.001, bs: 64, epochs: 30, MK-MMD loss weight: 1} \newline
\hbox{ResNet-50 $\rightarrow$ lr: 0.001, bs: 64, epochs: 30, MK-MMD loss weight: 0.3}

\item \uline{DANN}\newline
\hbox{AlexNet $\rightarrow$ lr: 0.001, bs: 64, epochs: 30, Domain Classifier loss weight: 0.1} \newline
\hbox{ResNet-50 $\rightarrow$ lr: 0.001, bs: 64, epochs: 30, Domain Classifier loss weight: 0.1}

\item \uline{AutoDIAL}\newline
AlexNet $\rightarrow$ lr: 0.0001, bs: 256, epochs: 30, Target loss weight ($\lambda$ in \eqref{lossautodial}): 0.4 \newline
ResNet-50 $\rightarrow$ lr: 0.001, bs: 256, epochs: 30, Target loss weight ($\lambda$ in \eqref{lossautodial}): 0.4

\item \uline{ADDA}\newline
ResNet-50 $\rightarrow$ lr: 0.001, bs: 128, epochs: 30, Mapping \& Adversarial loss weight: 1

\end{itemize}
\begin{table}
\centering
\begin{tabular}{lcc}
&\multicolumn{1}{c}{\textbf{AlexNet}}&\multicolumn{1}{c}{\textbf{ResnNet-50}}\\
Method & ROD $\rightarrow$ ARID & ROD $\rightarrow$ ARID\\ \hline\hline
Source only & 0.10 & 0.112\\
DAN &  0.144 &  0.203 \\
DANN &  0.148 & 0.182\\
AutoDIAL & \textbf{0.172} & 0.157 \\
ADDA & - & \textbf{0.228}
\end{tabular}
\caption{\label{tab:depth1}ROD $\rightarrow$ ARID Depth experiments of group (1).}\quad\begin{tabular}{lcc}
&\multicolumn{1}{c}{\textbf{AlexNet}}&\multicolumn{1}{c}{\textbf{ResnNet-50}}\\
Method & ROD $\rightarrow$ ARID & ROD $\rightarrow$ ARID\\ \hline\hline
Source only & 0.10 & 0.112\\
DAN &  0.144 &  0.207 \\
DANN &  0.145 & 0.19\\
AutoDIAL & \textbf{0.147} & 0.158 \\
ADDA & - & \textbf{0.209}
\end{tabular}
\caption{\label{tab:depth2}ROD $\rightarrow$ ARID Depth experiments of group (2).}
\end{table}
Referring to the set (1) of experiments, with AlexNet as starting net, we can observe that \ac{AutoDIAL} outperforms the other methods. Instead, the best result using ResNet-50 is reached with \ac{ADDA}. The primate of these two algorithms persists also for the second group of experiments although the highest accuracies are reached with the first one. Tables \ref{tab:depthbb1} and \ref{tab:depthbb2} show the best results obtained using depth informations as input data for the domain adaptation algorithms applied on the task \uline{\hbox{\ac{BigBIRD} $\rightarrow$ Active Vision}}. 
For this domain shift the set of parameters is the following:
\begin{itemize}
\item \uline{\ac{DAN}} \newline \hbox{AlexNet $\rightarrow$ lr: 0.001, bs: 64, epochs: 30, \ac{MK-MMD} loss weight: 1} \newline
\hbox{ResNet-50 $\rightarrow$ lr: 0.001, bs: 64, epochs: 30, \ac{MK-MMD} loss weight: 0.7}

\item \uline{\ac{DANN}}\newline
\hbox{AlexNet $\rightarrow$ lr: 0.0001, bs: 64, epochs: 30, Domain Classifier loss weight: 0.1} \newline
\hbox{ResNet-50 $\rightarrow$ lr: 0.001, bs: 128, epochs: 30, Domain Classifier loss weight: 0.1}

\item \uline{\ac{AutoDIAL}}\newline
\hbox{AlexNet $\rightarrow$ lr: 0.001, bs: 256, epochs: 30, Target loss weight ($\lambda$ in \eqref{lossautodial}): 0.1} \newline
ResNet-50 $\rightarrow$ lr: 0.001, bs: 256, epochs: 30, Target loss weight ($\lambda$ in \eqref{lossautodial}): 0.4

\item \uline{\ac{ADDA}}\newline
ResNet-50 $\rightarrow$ lr: 0.001, bs: 64, epochs: 30, Mapping \& Adversarial loss weight: 1
\end{itemize}

\begin{table}
\centering
\begin{tabular}{lcc}
&\multicolumn{1}{c}{\textbf{AlexNet}}&\multicolumn{1}{c}{\textbf{ResnNet-50}}\\
Method & BigBIRD $\rightarrow$ Active Vision & BigBIRD $\rightarrow$ Active Vision\\ \hline\hline
Source only & 0.041 & 0.055\\
DAN &  0.073 &  \textbf{0.11} \\
DANN &  0.067 & 0.094\\
AutoDIAL & \textbf{0.089} & 0.078 \\
ADDA & - & 0.097
\end{tabular}
\caption{\label{tab:depthbb1}BigBIRD $\rightarrow$ Active Vision Depth experiments of group (1).} \quad \begin{tabular}{lcc}
&\multicolumn{1}{c}{\textbf{AlexNet}}&\multicolumn{1}{c}{\textbf{ResnNet-50}}\\
Method & BigBIRD $\rightarrow$ Active Vision & BigBIRD $\rightarrow$ Active Vision\\ \hline\hline
Source only & 0.041 & 0.055\\
DAN &  0.08 &  \textbf{0.102} \\
DANN &  0.07 & 0.098\\
AutoDIAL & \textbf{0.084} & 0.101 \\
ADDA & - & 0.101
\end{tabular}
\caption{\label{tab:depthbb2}BigBIRD $\rightarrow$ Active Vision Depth experiments of group (2).}
\end{table}

Also in this case, for AlexNet, the best results are obtained using \ac{AutoDIAL} in both groups of experiments. With ResNet-50 as starting net, instead, the best accuracy is reached with \ac{DAN}. For depth, as we had expected also for RGB, the highest value is reached by the first group of experiments. Being depth informations less representative of the objects with respect to RGB data, obviously it is obtained a higher result using this last information. It is interesting now to see what happens if both inputs are used in the adaptation process.

\newpage
\subsection{RGB-D}

For the RGB-D experiments the high-level cue integration described in section 5.2 has been used. This approach is a starting point to see which is the best method to combine various image channels: in the deep domain adaptation scenario there are many other possible approaches to perform this combination, for example connecting both types of images (RGB and depth) at the \ac{CNN} network level as \cite{eitel2015multimodal} or \cite{hoffman2016cross}. The reason of the insertion of these experiments in the benchmark is to assess which is the improvement obtained adding informations, for instance about the shape, in an RGB object description. For RGB-D trials only experiments in which the whole target domain appears in both adaptation and testing phases are performed (group (1)). It was chosen to apply two domain adaptation algorithms for the task \hbox{\ac{ROD} $\rightarrow$ \ac{ARID}} using ResNet-50 as starting net: \ac{DAN} and \ac{DANN} (this last is the best for RGB experiments). It is good to remind that, for this set of experiments, depth images are colorized using the \hbox{Surface Normal ++} technique and, for \ac{ARID} dataset, only the depth images in which the number of non null pixels is greater than 75$\%$ are considered. Table \ref{tab:rgbdres} reports the results. 

\begin{table}[h]
\centering
\begin{tabular}{lc}
&\multicolumn{1}{c}{\textbf{ResnNet-50}}\\
Method & ROD $\rightarrow$ ARID\\ \hline\hline
Source only & 0.316\\
DAN &  0.439 \\
DANN &  \textbf{0.459} 
\end{tabular}
\caption{\label{tab:rgbdres}ROD $\rightarrow$ ARID RGB-D experiments of group (1).} 
\end{table}

\begin{figure}
\centering
\includegraphics[width=0.15\textwidth]{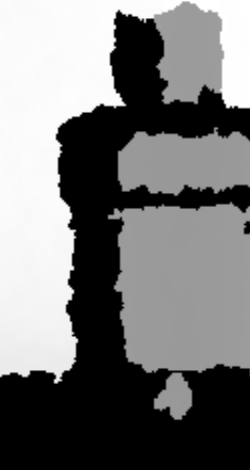}\quad\includegraphics[width=0.15\textwidth]{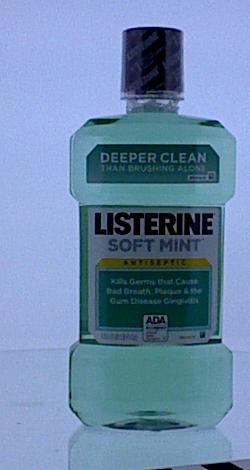}
\caption{\label{fig:listerinedepth}Example of the not perfect overlap between RGB and depth images in Big Berkeley Instance Recognition Dataset.}
\end{figure}

From the numbers in the table it can be notice that this type of combination of RGB and depth informations not produces improvements with respect to the results obtained with RGB only channel: the performance with \ac{DAN} decreases by $0.01\%$, instead, with \ac{DANN}, it remains the same. This could happen when the simplest way to combine data is used: the fact that the \ac{SVM} replaces the last fully connected layer in ResNet-50 led to a lower accuracy. For this reason it is important to consider a different baseline to evaluate the adaptation improvements (the one reported in Table \ref{tab:rgbdres}). As it can be notice the results for the adaptation task \hbox{\ac{BigBIRD} $\rightarrow$ Active Vision} are missing. This is due to the not perfect overlap between RGB and depth images. The segmentation masks used for the crops (see paragraph \textit{Datasets} of Section 6.2) are not well calibrated with depth images resulting in a not precise object bounding box. Figure \ref{fig:listerinedepth} shows an explicit example in which the size of the cropped image is the same but in the depth one the object is not well centered.

\newpage
\chapter{Conclusions}

To summarize, the following conclusions were reached: (i) Table \ref{tab:improvement} shows that the DA methods do not work on depth with the same effectiveness with which they work on RGB data, (ii) always from Table \ref{tab:improvement} it can be seen that the high level integration method used to combine RGB and depth produces a good improvement with respect to its baseline, but not with respect the accuracy obtained with the same experiment using only RGB as input data. (i) and (ii) prove that how deal with depth informations is still an open problem that needs specific research efforts to be used properly. (iii) \ac{AutoDIAL} outperforms in every experiment (except one) which has AlexNet as starting network (see the first two columns of Table \ref{tab:comment1}), (iv) \ac{DANN} outperforms for the greatest part of the experiments which have ResNet-50 as starting network (see Table \ref{tab:comment1}), (v) the accuracy obtained applying an adaptation algorithm on the task \hbox{\ac{WOD} $\rightarrow$ \ac{ARID}} is higher with respect to the one obtained applying the same algorithm to the shift \hbox{\ac{ROD} $\rightarrow$ \ac{ARID}}, (vi) the group of experiments that reaches an higher accuracy, in the case of RGB input data, is the one in which two different subsets of the target dataset are used for the adaptation and testing phases (group (2)) (see first row of Table \ref{tab:comment2}), instead,(vii) for all experiments with only depth as input data the best accuracy is reached with the group that uses the same dataset for both adaptation and testing phases (group (1)) (see second row Table \ref{tab:comment2}), (viii) the model trained with \ac{ROD} depth images produces a better performance if it is tested on the \ac{ARID} depth images in which the objects are within a meter from the camera (see Table \ref{tab:comment3}).

In this work it has been presented a new benchmark valuables for the robot vision community in the object recognition field. In particular, the core of its contribution is in the investigation of domain adaptation with multimodal input data. The novelty is mostly in the use of depth: a lot of domain adaptation studies were conducted on RGB images but not on depth and even less on RGB combined with depth. Some off-the-shelf unsupervised domain adaptation algorithms, together with datasets suited for robotic purposes, have been used.
The experiments conducted shows that domain alignment, especially in the depth space, is still an open problem that needs more research efforts to be used properly.
\begin{table}[H]
\begin{tabular}{lcccc}
&\multicolumn{2}{c}{\textbf{AlexNet}}&\multicolumn{2}{c}{\textbf{ResNet-50}}\\
Group of \\ experiments & ROD $\rightarrow$ ARID & BB $\rightarrow$ AV & ROD $\rightarrow$ ARID & BB $\rightarrow$ AV\\ \hline\hline
RGB (1) &  \textbf{+8.7\%} (37.8\%) &  +25.7\% (52.1\%) & +12.2\% (45.9\%) &  +17.6\% (53.7\%) \\
RGB (2) &  +6.5\% (35.6\%) & \textbf{+29.4\%} (55.8\%) & +13\% (46.7\%) & \textbf{+18.1\%} (54.2\%)\\
Depth (1) & +7.2\% (17.2\%) & +4.8\% (8.9\%) & +11.6\% (22.8\%) & +5.5\% (11\%) \\
Depth (2) & +4.7\% (14.7\%) & +4.3\% (8.4\%) & +9.7\% (20.9\%) & +4.7\% (10.2\%) \\
RGB-D & & & \textbf{+14.3}\% (45.9\%) 
\end{tabular}
\caption{\label{tab:improvement} Best improvement reached through domain adaptation algorithms for each RGB only, depth only and RGB-D task. In brackets also the accuracy reached is shown.}
\end{table}
\begin{table}[H]
\centering
\begin{tabular}{lcccc}
&\multicolumn{2}{c}{\textbf{AlexNet}}&\multicolumn{2}{c}{\textbf{ResNet-50}}\\
Group of experiments & ROD $\rightarrow$ ARID & BB $\rightarrow$ AV & ROD $\rightarrow$ ARID & BB $\rightarrow$ AV\\ \hline\hline
RGB (1) &  \textbf{AutoDIAL} &  \textbf{AutoDIAL} & \textbf{DANN} & \textbf{DANN} \\
RGB (2) &  DANN & \textbf{AutoDIAL} & AutoDIAL & \textbf{DANN} \\
Depth (1) & \textbf{AutoDIAL} & \textbf{AutoDIAL} & ADDA & DAN\\
Depth (2) & \textbf{AutoDIAL} & \textbf{AutoDIAL} & ADDA & DAN
\end{tabular}
\caption{\label{tab:comment1} Best algorithm for each RGB only and depth only experiment.}
\end{table}
\begin{table}[H]
\centering
\begin{tabular}{lcccc}
&\multicolumn{2}{c}{\textbf{AlexNet}}&\multicolumn{2}{c}{\textbf{ResNet-50}}\\
Input modality & ROD $\rightarrow$ ARID & BB $\rightarrow$ AV & ROD $\rightarrow$ ARID & BB $\rightarrow$ AV\\ \hline\hline
RGB &  (1) &  (2) & (2) & (2) \\
Depth & (1) & (1) & (1) & (1)
\end{tabular}
\caption{\label{tab:comment2} Group of experiments with which the best result is obtained.}
\end{table}
\begin{table}[H]
\centering
\begin{tabular}{lccccccccc}
\textbf{Range} (mm) & 0-1000 & 0-1200 & 0-1400 & 0-1600 & 0-1800 & 0-2000 & 0-2200 & 0-2400 & All\\
\textbf{Accuracy} &  \textbf{0.166} & 0.163 & 0.142 & 0.127 & 0.116 & 0.108 & 0.102 & 0.0996 & 0.0995
\end{tabular}
\caption{\label{tab:comment3} Accuracies obtained using as test set progressive subsets (in terms of camera - object distance) of Autonomous Robot Indoor Dataset.}
\end{table}

\newpage
\addcontentsline{toc}{chapter}{\bibname}
\printbibliography

\end{document}